\documentclass[a4paper,11pt]{article}

\usepackage[top=2.54cm,bottom=2.54cm, left=2.54cm, right=2.54cm]{geometry}

\usepackage{amsmath}
\usepackage{amsfonts}
\usepackage{amssymb}
\usepackage{graphicx}
\usepackage{hyperref}
\usepackage{bm}
\newtheorem{lem}{Lemma}[section]
\usepackage{enumerate}
\usepackage{color}
\usepackage[sort,comma,numbers]{natbib}
\usepackage{float}
\begin{document}

\begin{center}
\begin{Large}
A folded model for compositional data analysis \\
\end{Large}
\vskip 0.5cm
Michail Tsagris$^1$ and Connie Stewart$^2$ \\
\bigskip
$^1$ Department of Economics, University of Crete, Rethymnon, Greece, mtsagris@uoc.gr  \\
$^2$ Department of Mathematics and Statistics, University of New Brunswick, Saint John, New Brunswick Canada,
connie.stewart@unb.ca
\end{center}

\begin{center}
{\bf Abstract}
\end{center}
A folded type model is developed for analyzing compositional data. The proposed model involves an extension of the $\alpha$-transformation for compositional data and provides a new and flexible class of distributions for modeling data defined on the simplex sample space. Despite its rather seemingly complex structure, employment of the EM algorithm guarantees efficient parameter estimation. The model is validated through simulation studies and examples which illustrate that the proposed model performs better in terms of capturing the data structure, when compared to the popular logistic normal distribution, and can be advantageous over a similar model without folding. 

\section{Introduction}
Compositional data are positive multivariate data which sum to the same constant, usually $1$. In this case, their sample space is the standard simplex
\begin{eqnarray} \label{simplex}
\mathbb{S}^{D-1}=\left\lbrace(x_1,...,x_D)^T \bigg\vert x_i \geq 0,\sum_{i=1}^Dx_i=1\right\rbrace, 
\end{eqnarray}
where $D$ denotes the number of variables (better known as components). 

Compositional data are met in many different scientific fields. In sedimentology, for example, samples were taken from an Arctic lake and their composition of water, clay and sand were the quantities of interest. Data from oceanography studies involving Foraminiferal (a marine plankton species) compositions at $30$ different sea depths from oceanography were analyzed in Aitchison (2003, pg 399). Schnute \& Haigh (2007) analyzed marine compositional data through catch curve models for a quillback rockfish (\textit{Sebastes maliger}) population. 
In hydrochemistry, Otero et al. (2005) used regression analysis to draw conclusions about anthropogenic and geological pollution sources of rivers in Spain. Stewart \& Field (2011) modeled compositional diet estimates with an abundance of zero values obtained through quantitative fatty acid signature analysis. In another biological setting, Ghosh \& Chakrabarti (2009) were interested in the classification of immunohistochemical data. Other application areas of compositional data analysis include archaeometry (Baxter et al., 2005), where the composition of ancient glasses, for instance, is of interest, and economics (Fry, Fry, \& McLaren, 2000), where the focus is on the percentage of the household expenditure allocated to different products. Compositional data are also met in political science (Katz \& King, 1999) for modeling electoral data and in forensic science where the compositions of forensic glasses are compared and classified (Neocleous, Aitken \& Zadora, 2011). In demography, compositional data are met in multiple-decrement life tables and the mortality rates amongst age groups are modeled (Oeppen, 2008). In a study of the brain, Prados et al. (2010) evaluated the diffusion anisotropy from diffusion tensor imaging using new measures derived from compositional data distances. Some recent areas of application include bioinformatics and specifically microbiome data analysis (Xia et al., 2013; Chen \& Li, 2016; Shi, Zhang \& Li, 2016). These examples illustrate the breadth of compositional data analysis applications and consequently the need for parametric models defined on the simplex.

The Dirichlet distribution is a natural distribution for such data due to its support being the simplex space. However, it has long been recognized that this distribution is not, statistically, rich and flexible enough to capture many types of variabilities (especially curvature) of compositional data. For this reason, a variety of transformations have been proposed that map the data outside of the simplex. In Aitchison (1982) the log-ratio transformation approach was developed, and later the so called isometric log-ratio transformation methodology which was first proposed in Aitchison (2003, pg 90) and examined in detail by Egozcue et al. (2003). More recently, Tsagris, Preston \& Wood (2011) suggested the $\alpha$-transformation which includes the isometric transformation as a special case. The $\alpha$-transformation is a Box-Cox type transformation and has been successfully applied in regression analysis (Tsagris, 2015) and classification settings (Tsagris, 2014; Tsagris, Preston \& Wood, 2016).

Regardless of the transformation chosen, the usual approach for modeling compositional data is to assume that the transformed data are multivariate normally distributed.  While the $\alpha$-transformation offers flexibility, a disadvantage of this transformation is that it maps the compositional data from the simplex ($\mathbb{S}^{D-1}$) to a subset of $\mathbb{R}^{D-1}$, and not $\mathbb{R}^{D-1}$ itself on which the multivariate normal density is defined\footnote{In the univariate case, for example, imagine fitting a normal (and not a truncated normal) to a limited space of its support.}. An improvement to this method can be obtained by using a folded model procedure, similar to the approach used in Scealy \& Welsh (2014) and to the folded normal distribution in $\mathbb{R}$ employed in Leone, Nelson \& Nottingham (1961) and Johnson (1962). The folded normal distribution in $\mathbb{R}$, for example, corresponds to taking the absolute value of a normal random variable and essentially ``folding'' the negative values to the positive side of the distribution. The model we propose here works in a similar fashion where values outside the simplex are mapped inside of it via a folding type transformation. An advantage of this approach over the aforementioned log-ratio methodology is that it allows one to fit any suitable multivariate distribution on $\mathbb{S}^{D-1}$ through the parameter $\alpha$.

The paper is structured as follows. In Section \ref{foldingtech} we describe the folding approach and in Section \ref{insideout} we introduce the $\alpha$-folded multivariate normal model for compositional data, along with the EM algorithm for maximum likelihood estimation, an algorithm for generating data from the proposed folded distribution and inference for the transformation parameter $\alpha$. Four data sets are analyzed in Section \ref{data} that illustrate both the advantages and limitations of the proposed model. Finally, simulation studies are provided in Section \ref{sim} to assess the estimation accuracy of the parameters and computational burden of the algorithm, and concluding remarks may be found in Section \ref{conclusion}.
 
\section{The $\alpha$ Folding Technique} 
\label{foldingtech}
\subsection{The $\alpha$-Transformation}
For a composition $\mathbf{x} \in \mathbb{S}^{D-1}$, the centered log-ratio transformation is defined in Aitchison (1983) as 

\begin{eqnarray} \label{clr}
\mathbf{w}^0(\mathbf{x})=\left ( \log\left ({\frac{x_1}{\prod_{j=1}^Dx_j^{1/D}}} \right ),\ldots, \log\left ({\frac{x_D}{\prod_{j=1}^Dx_j^{1/D}}} \right ) \right ).
\end{eqnarray}

The sample space of Equation (\ref{clr}) is the set
\begin{eqnarray} \label{Qd}
\mathbb{Q}_0^{D-1}=\left\lbrace \left(w^0_1, \ldots, w^0_D \right)^T: \sum_{i=1}^Dw^0_i=0 \right\rbrace
\end{eqnarray}
which is a subset of $\mathbb{R}^{D-1}$. Note that the zero sum constraint in Equation (\ref{Qd}) is an obvious drawback of this transformation as it can lead to singularity issues. In order to remove the redundant dimension imposed by this constraint, one can apply the isometric log-ratio transformation 
\begin{eqnarray} \label{ilr}
{\bf z}_0(\mathbf{x})=\mathbf{H}\mathbf{w}_0(\mathbf{x}),
\end{eqnarray}
where ${\bf H}$ is the Helmert matrix (Lancaster, 1965) (an orthonormal $D\times D$ matrix) after deletion of the first row. This matrix is referred to as the Helmert sub-matrix\footnote{The Helmert sub-matrix is a standard orthogonal matrix in shape analysis used to overcome singularity problems (Dryden \& Mardia, 1998; Le \& Small, 1999).} and its structure and components are specified in the Appendix. Left multiplication by the Helmert sub-matrix maps the data onto $\mathbb{R}^{D-1}$ thus, in effect, removing the zero sum constraint. 

Tsagris, Preston \& Wood (2011) developed the $\alpha$-transformation as a more general transformation than that in Equation (\ref{ilr}). Let 
\begin{eqnarray} 
\label{stayalpha}
{\bf u}_{\alpha}(\mathbf{x})=\left( \frac{x_1^{\alpha}}{\sum_{j=1}^Dx_j^{\alpha}}, \ldots, \frac{x_D^{\alpha}}{\sum_{j=1}^Dx_j^{\alpha}} \right)^T
\end{eqnarray}
denote the power transformation for compositional data as defined by Aitchison (2003). In a manner analogous to Equations (\ref{clr}-\ref{ilr}), first define
\begin{eqnarray} \label{alef}
{\bf w}_{\alpha}(\textbf{x})=\frac{D{\bf u}_{\alpha}-1}{\alpha}.
\end{eqnarray}
The sample space of Equation (\ref{alef}) is then the set
\begin{eqnarray}  \label{Qad}
\mathbb{Q}_{\alpha}^{D-1}=\left\lbrace \left(w_{1, \alpha}, \ldots, w_{D, \alpha} \right)^T:\frac{-1}{\alpha} \leq w_{i, \alpha} \leq \frac{D-1}{\alpha},\sum_{i=1}^Dw_{i,\alpha}=0 \right\rbrace.
\end{eqnarray}
Note that the inverse of Equation (\ref{alef}) is as follows  
\begin{equation}
\label{winv}
\mathbf{x} =\mathbf{w}^{-1}_{\alpha}(\mathbf{m}) = \left (\frac{(1+\alpha m_1)^{1/\alpha}}{\sum_{j=1}^D (1+\alpha m_j)^{1/\alpha}},\ldots,\frac{(1+\alpha m_D)^{1/\alpha}}{\sum_{j=1}^D (1+\alpha m_j)^{1/\alpha}} \right )
\end{equation}
for $\mathbf{m} \in \mathbb{Q}_{\alpha}^{D-1}$. As $\alpha \rightarrow 0$ Equation (\ref{alef}) converges to Equation (\ref{clr}) and Equation (\ref{winv}) becomes
\begin{equation}
\label{winv0}
\mathbf{x} =\mathbf{w}^{-1}_0(\mathbf{m}) = \left (\frac{e^ {m_1}}{\sum_{j=1}^D e^{m_j}},\ldots,\frac{e^ {m_D}}{\sum_{j=1}^D e^{m_j}} \right ).
\end{equation}

Finally, the $\alpha$-transformation is defined as
\begin{eqnarray} \label{alpha}
{\bf z}_{\alpha}(\mathbf{x})={\bf H}{\bf w}_{\alpha}(\mathbf{x}).
\end{eqnarray}
The transformation in Equation (\ref{alpha}) is a one-to-one transformation which maps data inside the simplex onto a subset of $\mathbb{R}^{D-1}$ and vice versa for 
$\alpha \neq 0$. The corresponding sample space of Equation (\ref{alpha}) is 
\begin{eqnarray} \label{Ad}
\mathbb{A}_{\alpha}^{D-1}=\left\lbrace{\bf H}{\bf w}_{\alpha} \bigg | -\frac{1}{\alpha} \leq w_{i,\alpha} \leq \frac{D-1}{\alpha},\sum_{i=1}^Dw_{i, \alpha}=0 \right\rbrace.
\end{eqnarray}
For $\mathbf{y} = \mathbf{z}_\alpha(\mathbf{x})$ , the inverse transformation from $\mathbb{A}_{\alpha}^{D-1}$ to $\mathbb{S}^{D-1}$ is $\mathbf{z}_\alpha^{-1}(\mathbf{y}) = \mathbf{w}^{-1}_\alpha(\mathbf{H}^T\mathbf{y})$ where $\mathbf{w}^{-1}(\cdot)$ is given in Equation (\ref{winv}). Note that vectors in $\mathbb{A}_{\alpha}^{D-1}$ are not subject to the sum to zero constraint and that $\lim_{\alpha \rightarrow 0}\mathbb{A}_{\alpha}^{D-1} \rightarrow \mathbb{R}^{D-1}$.

For convenience purposes we allow $\alpha$ to lie within $\left[-1,1\right]$. From Equations (\ref{stayalpha}) and (\ref{alef}), when $\alpha=1$, the simplex is linearly expanded as the values of the components are simply multiplied by a scalar and then centered. When $\alpha=-1$, the inverse of the values of the components are multiplied by a scalar and then centered.

If we assume that the $\alpha$-transformed data (for any value of $\alpha \in \left[-1,1\right]$) follow a multivariate normal distribution, then a way to choose the value of $\alpha$ is via maximum likelihood estimation (MLE). However, given that the multivariate normal distribution is defined on $\mathbb{R}^{D-1}$ and that $\mathbb{A}_{\alpha}^{D-1} \subseteq \mathbb{R}^{D-1}$, this approach might neglect an important amount of volume (probability) of the multivariate normal. The same problem arose in Scealy \& Welsh (2011b), who then developed the folded Kent distribution (Scealy \& Welsh, 2014) and we propose a similar solution here.  

\subsection{The $\alpha$-Folding Transformation} \label{foldedjacobian}

Let $\mathbf{y} = \mathbf{z}_\alpha(\mathbf{x})$ in Equation (\ref{alpha}) for $\mathbf{x} \in \mathbb{S}^{D-1}$ and some value of $\alpha$. The inverse transformation $\mathbf{z}_{\alpha}^{-1}(\mathbf{y})$ provides a transformation from $\mathbf{y} \in \mathbb{A}_{\alpha}^{D-1}$ to $\mathbb{S}^{D-1}$. Now suppose we have a point ${\bf y} \in \mathbb{R}^{D-1}\setminus\mathbb{A}_{\alpha}^{D-1}$ (that is, outside of $\mathbb{A}_{\alpha}^{D-1}$) and we want to map it inside the simplex $\mathbb{S}^{D-1}$. As proven in the Appendix, the following transformation maps $\mathbf{y}$ to $\mathbb{S}^{D-1}$
\begin{eqnarray} \label{foldin}
{\bf x}={\bf w}_{\alpha}^{-1}\left( \frac{ \mathbf{H}^T\mathbf{y}}{ q^{*2}_\alpha(\mathbf{y}) }\right),
\end{eqnarray}
where ${\bf w}_{\alpha}^{-1}\left(.\right)$ is defined in Equation (\ref{winv}), $q_\alpha^{*}(\mathbf{y})=\alpha\min \left \lbrace \mathbf{H}^T\mathbf{y} \right\rbrace$ and $\min \left \lbrace \mathbf{H}^T\mathbf{y} \right\rbrace$ refers to the minimum component of the vector $\mathbf{H}^T\mathbf{y}$. 

Note that for ${\bf y} \in \mathbb{R}^{D-1}\setminus\mathbb{A}_{\alpha}^{D-1}$ the inverse of Equation (\ref{foldin}) is the $D-1$ dimensional vector
\begin{eqnarray} \label{foldout}
{\bf y} =\frac{1}{w_\alpha^{*2}(\mathbf{x})}{\bf H}\mathbf{w}_\alpha(\mathbf{x})=\frac{1}{w_\alpha^{*2}(\mathbf{x})}\mathbf{z}_\alpha(\mathbf{x}),
\end{eqnarray}
where $w^{*}_\alpha(\mathbf{x})=\alpha \min \left\lbrace \mathbf{w}_\alpha(\mathbf{x}) \right\rbrace $ and $\min \left\lbrace \mathbf{w}_\alpha(\mathbf{x}) \right\rbrace$ is the minimum component of the vector ${\bf w}_{\alpha}(\mathbf{x})$ defined in Equation (\ref{alef}).

The square term in the exponent of the denominator in (\ref{foldin}) was chosen for convenience purposes. The square could alternatively be substituted by any other power say $\beta$. However, this generalization would make the model more complex and difficult to work with as there would be an additional parameter to estimate.  In the Bayesian stance this would be easier to solve, yet less efficient.

In summary, we propose the following folding transformation from $\mathbf{y}\in \mathbb{R}^{D-1}$ to $\mathbb{S}^{D-1}$

\begin{equation} 
\label{trans}
\mathbf{x} = \left \{ \begin{array}{ll}
 g_0^\alpha(\mathbf{y}) & \mathrm{if\ } \mathbf{y} \in \mathbb{A}_\alpha^{D-1} \\
 g_1^\alpha(\mathbf{y}) & \mathrm{if\ } \mathbf{y} \in \mathbb{R}^{D-1}\setminus\mathbb{A}_\alpha^{D-1}
\end{array} \right.
\end{equation}
where $g_0^\alpha(\mathbf{y}) = \mathbf{z}_\alpha^{-1}(\mathbf{y})=\mathbf{w}^{-1}_\alpha( \mathbf{H}^T\mathbf{y})$ and $g_1^\alpha(\mathbf{y}) = {\bf w}_{\alpha}^{-1}\left( \frac{ 
 \mathbf{H}^T\mathbf{y} }{ q^{*2}_\alpha(\mathbf{y}) }\right)$. 
\ \\
\noindent

\section{The $\alpha$-Folded Multivariate Normal Distribution} \label{insideout}

\subsection{Definition}
\label{definition}
The derivation of the $\alpha$-folded multivariate normal distribution is carried out by first assuming that $\mathbf{y}$ is multivariate normally distributed with parameters $\pmb{\mu}_\alpha$ and $\pmb{\Sigma}_\alpha$, and subsequently considering the distribution of $\mathbf{x}$ defined in Equation (\ref{trans}).  The basis of this approach was used in Leone (1961) to derive the univariate folded normal distribution.

If $\mathbf{y}\sim N^{D-1}\left(\pmb{\mu}_\alpha,\pmb{\Sigma}_\alpha \right )$, then the density of $\mathbf{y}$ is given by 
\begin{equation}
\label{MVN}
f_\mathbf{y}(\mathbf{y}) = \frac{1}{|2\pi \pmb{\Sigma}_\alpha|^{1/2}}\exp{\left[-\frac{1}{2}\left(\mathbf{y}-\pmb{\mu}_{\alpha}\right)^T\pmb{\Sigma}^{-1}_{\alpha}\left(\mathbf{y}-\pmb{\mu}_{\alpha}\right)\right]},
\end{equation}
and we can apply standard statistical techniques for finding the distribution of a transformation of a random vector to derive the distribution of $\mathbf{x}$, recognizing that in our case we require the distribution of two transformations, namely $g_0^\alpha(\mathbf{y})$ and $g_1^\alpha(\mathbf{y})$.  It is straightforward to derive the densities of $g_0^\alpha(\mathbf{y})$ and $g_1^\alpha(\mathbf{y})$ since the Jacobians of the transformations are known, and they are given in the following two Lemmas with corresponding proofs in the Appendix.
\begin{lem} 
\label{j0}
The Jacobian of $\mathbf{z}_\alpha(\mathbf{x})$ is
\begin{eqnarray*}
\left|J^0_{\alpha} \right|
=D^{D-1+\frac{1}{2}}\prod_{i=1}^D\frac{x_i^{\alpha-1}}{\sum_{j=1}^D x_j^{\alpha}}
\end{eqnarray*}
\end{lem}
\begin{lem}
The Jacobian of $\frac{\mathbf{z}_\alpha(\mathbf{x})}{w_\alpha^{*2}(\mathbf{x})}$ is 
\begin{eqnarray*}
\left|J^1_{\alpha}\right|=D^{D-1+\frac{1}{2}}\prod_{i=1}^D\frac{x_i^{\alpha-1}}{\sum_{j=1}^D x_j^{\alpha}} \times \left(\frac{1}{\alpha w^*_\alpha(\mathbf{x})}\right)^{2(D-1)}.
\end{eqnarray*}

\end{lem}

Noting that the inverse of $g_0^\alpha(\mathbf{y})$ is $\mathbf{z}_\alpha(\mathbf{x})$ and that of $g_1^\alpha(\mathbf{y})$ is $\frac{\mathbf{z}_\alpha(\mathbf{x})}{w_\alpha^{*2}(\mathbf{x})}$, the densities based on the transformations $g_0^\alpha$ and $g_1^\alpha$ are given in Equations (\ref{f0}) and (\ref{f1}) respectively as
\begin{equation}
\label{f0}
 f_{\mathbf{x}_0}\left({\bf x}|\alpha \right)=\frac{\left|J^0_{\alpha}\right|}{{\left|2\pi\pmb{\Sigma}_{\alpha}\right|^{1/2}}}
\exp{\left[-\frac{1}{2}\left(\mathbf{z}_\alpha(\mathbf{x})-\pmb{\mu}_{\alpha}\right)^T\pmb{\Sigma}^{-1}_{\alpha}\left(\mathbf{z}_\alpha(\mathbf{x})-\pmb{\mu}_{\alpha}\right)\right]} 
\end{equation}
\begin{equation}
\label{f1}
f_{\mathbf{x}_1}\left({\bf x}|\alpha \right)=\frac{\left|J^1_{\alpha}\right| }
{ \left|2\pi\pmb{\Sigma}_{\alpha}\right|^{1/2} }
\exp{ \left[-\frac{1}{2}\left(\frac{\mathbf{z}_\alpha(\mathbf{x})}{w^{*2}_\alpha(\mathbf{x})} -\pmb{\mu}_{\alpha}\right)^T\pmb{\Sigma}^{-1}_{\alpha}
  \left(\frac{\mathbf{z}_\alpha(\mathbf{x})}{w_\alpha^{*2}(\mathbf{x})} -\pmb{\mu}_{\alpha}\right)
  \right]}.
\end{equation}

If we let $p$ denote the probability that $\mathbf{y} \in \mathbb{A}^{D-1}$, the distribution of $\mathbf{x}\in \mathbb{S}^{D-1}$ can be written as
\begin{equation} 
\label{mixture}
f_\mathbf{x}\left({\mathbf{x}}|\alpha,p \right) = p f_{\mathbf{x}_0}\left({\bf x}|\alpha \right)+ \left(1-p\right) f_{\mathbf{x}_1}\left({\bf x}|\alpha \right).
\end{equation}
In words, $1-p$ is the probability that $\mathbf{y}$ needs to be folded into the simplex through the transformation $g_1^\alpha$ in Equation (\ref{trans}).
Substituting the density functions in Equations (\ref{f0}) and (\ref{f1}) into Equation (\ref{mixture}), it follows that




\begin{eqnarray}
\label{foldednormal}
f_\mathbf{x}\left({\mathbf{x}}|\alpha,p,\pmb{\mu}_{\alpha},\pmb{\Sigma}_{\alpha} \right) & = &  
p \frac{\left|J^0_{\alpha}\right|}{{\left|2\pi\pmb{\Sigma}_{\alpha}\right|^{1/2}}}
\exp{\left[-\frac{1}{2}\left(z_\alpha(\mathbf{x})-\pmb{\mu}_{\alpha}\right)^T\pmb{\Sigma}^{-1}_{\alpha}\left(z_\alpha(\mathbf{x})-\pmb{\mu}_{\alpha}\right)\right]} 
\\ \nonumber
& + & \left(1-p\right) 
\frac{\left|J^1_{\alpha}\right| }
{ \left|2\pi\pmb{\Sigma}_{\alpha}\right|^{1/2} }
\exp{ \left[-\frac{1}{2}\left(\frac{z_\alpha(\mathbf{x})}{w^{*2}_\alpha(\mathbf{x})} -\pmb{\mu}_{\alpha}\right)^T\pmb{\Sigma}^{-1}_{\alpha}
  \left(\frac{z_\alpha(\mathbf{x})}{w_\alpha^{*2}(\mathbf{x})} -\pmb{\mu}_{\alpha}\right)
  \right]},
\end{eqnarray}

where ${\bf x} \in \mathbb{S}^{D-1}$, $\alpha \in [-1,1]$, $0\leq p \leq 1$.  We will refer to the density in Equation (\ref{foldednormal}) as the $\alpha$-folded multivariate normal distribution.

Although we have indicated that ${\bf x} \in \mathbb{S}^{D-1}$, it is important to note that boundaries on the simplex (that is, zero values) are not allowed due to the product in Lemma \ref{j0}. This potential limitation also occurs with conventional transformations for compositional data analysis as well, including the isometric transformation in Equation (\ref{ilr}).  Recent approaches to handle zero values include mapping the data onto the hyper-sphere (Scealy and Welsh, 2011b, Scealy and Welsh), assuming latent variables (Butler and Glasbey, 2008) or conditioning on the zero values (Stewart and Field 2011; Tsagris and Stewart, 2018).

A few special cases are worthy of mention. First, when $p=1$, the second term in Equation (\ref{foldednormal}) vanishes and we end up with the $\alpha$-normal distribution (Tsagris, Preston and Wood, 2011). When $\alpha=1$, $\left|J^0_{\alpha} \right|=D^{D-1+\frac{1}{2}}$. Finally, when $\alpha \rightarrow 0$, Equation (\ref{foldednormal}) reduces to the multivariate logistic normal on $\mathbb{S}^d$ (Aitchison, 2003). 
\begin{eqnarray} \label{ilr density}
f_\mathbf{x}\left({\bf x}\right) &=& \frac{1}{\left|2\pi {\bf \Sigma}\right|^{1/2}} \mbox{exp}\left[ 
-\frac{1}{2}\left(z_0(\mathbf{x})-\pmb{\mu}_0\right)^T {\bf \Sigma}_0^{-1}\left(z_0(\mathbf{x})-\pmb{\mu}_0\right)\right]\prod_{i=1}^Dx_i^{-1},
\end{eqnarray} 
where $z_0(\mathbf{x})$ is defined in Equation (\ref{ilr}). 

We can mimic Aitchison's (2003) Definition 6.2 of the additive logistic normal distribution and similarly define the $\alpha$-folded multivariate normal distribution by a transformation to multivariate normality. Specifically, a D-part composition ${\bf x}$ is said to follow the $\alpha$-folded multivariate normal distribution  if $\mathbf{y}$ follows a multivariate normal distribution in $\mathbb{R}^d$ where
\begin{equation} 
\label{invtrans}
\mathbf{y} = \left \{ \begin{array}{ll}
{\bf z}_{\alpha}(\mathbf{x}) & \mathrm{with\  probability\ } p \\
 \frac{{\bf z}_{\alpha}(\mathbf{x})}{w_\alpha^{*2}(\mathbf{x})}, & \mathrm{with\ probability} 1-p,
\end{array} \right.
\end{equation}
$w^{*}_\alpha(\mathbf{x})= \alpha \min \left\lbrace  \mathbf{w}_{\alpha}(\mathbf{x}) \right\rbrace$ and ${\bf z}_{\alpha}(\mathbf{x})$ is defined in Equation (\ref{alpha}).  


Unlike in the case of log-ratio transformations to normality, numerical optimization is required for obtaining the maximum likelihood estimates of the parameters in the $\alpha$-folded multivariate normal distribution and the procedure is fully described in Subsection \ref{mle}. 

The $\alpha$-folded multivariate normal distribution resolves the problem associated with the assumption of multivariate normality of the $\alpha$-transformed data (Tsagris, Preston \& Wood, 2011); the ignored probability left outside the simplex due to the sample space of the $\alpha$-transformation being a subset of $\mathbb{R}^{D-1}$. With this distribution, any ignored probability is folded back onto the simplex, and hence the density has the form of Equation (\ref{foldednormal}). 

\subsection{Contour plots of the $\alpha$-folded bivariate normal distribution}
\label{contours}

To help visualize the $\alpha$-folded multivariate normal distribution, we consider the two-dimensional case and compare the contours of the bivariate normal distribution with and without folding. In particular, we examine contours of the normal distribution in $\mathbb{R}^2$ and compare these to the contours of the folded model in Equation (\ref{foldednormal}) plotted on the simplex, with $\alpha=1$.  We consider two settings in which $\pmb{\mu}=\left(0.561, 0.547 \right)^T$ in both cases, but the covariances matrices differ as follows
\begin{eqnarray} \label{Sigma}
\pmb{\Sigma}_1=\left(
\begin{array}{cc} 
0.5  & 0.25   \\
0.25 & 0.35   \\
\end{array} \right) 
\ \text{and} \ \ 
\pmb{\Sigma}_2=\left(
\begin{array}{cc} 
2.5  & 1.25   \\
1.25 & 1.75   \\
\end{array} \right). 
\end{eqnarray}

The mean vector was selected by applying the $\alpha$-transformation in Equation (\ref{alpha}) with $\alpha=1$ to a sub-composition formed by the first three components of the Hongite data (an artificial data set used by Aitchison, 2003). The elements of the first covariance matrix ($\pmb{\Sigma}_1$) were chosen so that the correlation was positive, whereas the second covariance matrix is such that $\pmb{\Sigma}_2 = 5\pmb{\Sigma}_1$.  

\begin{figure}[!h]
\centering
\begin{tabular}{cc}
\includegraphics[scale=0.3,trim=0 0 0 0]{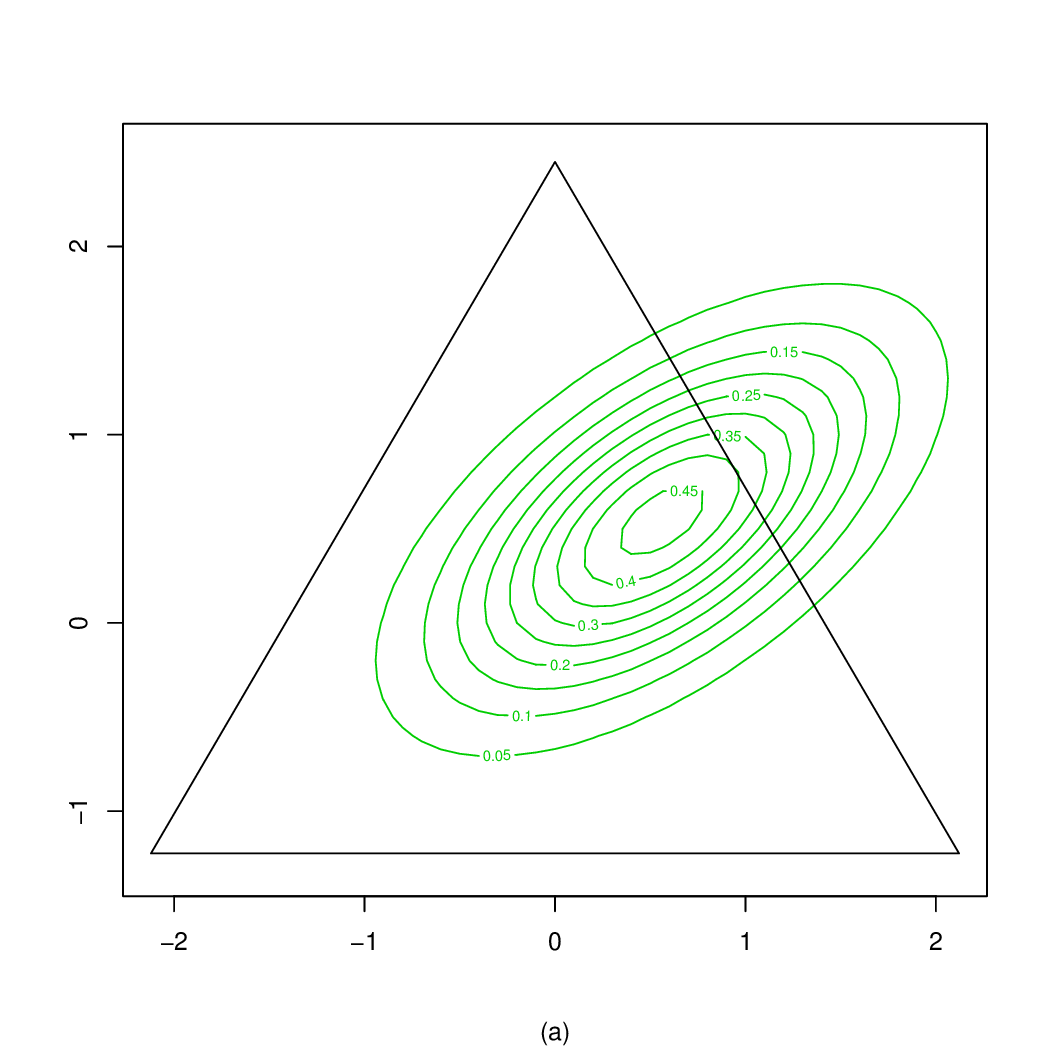} &
\includegraphics[scale=0.3,trim=0 0 0 0]{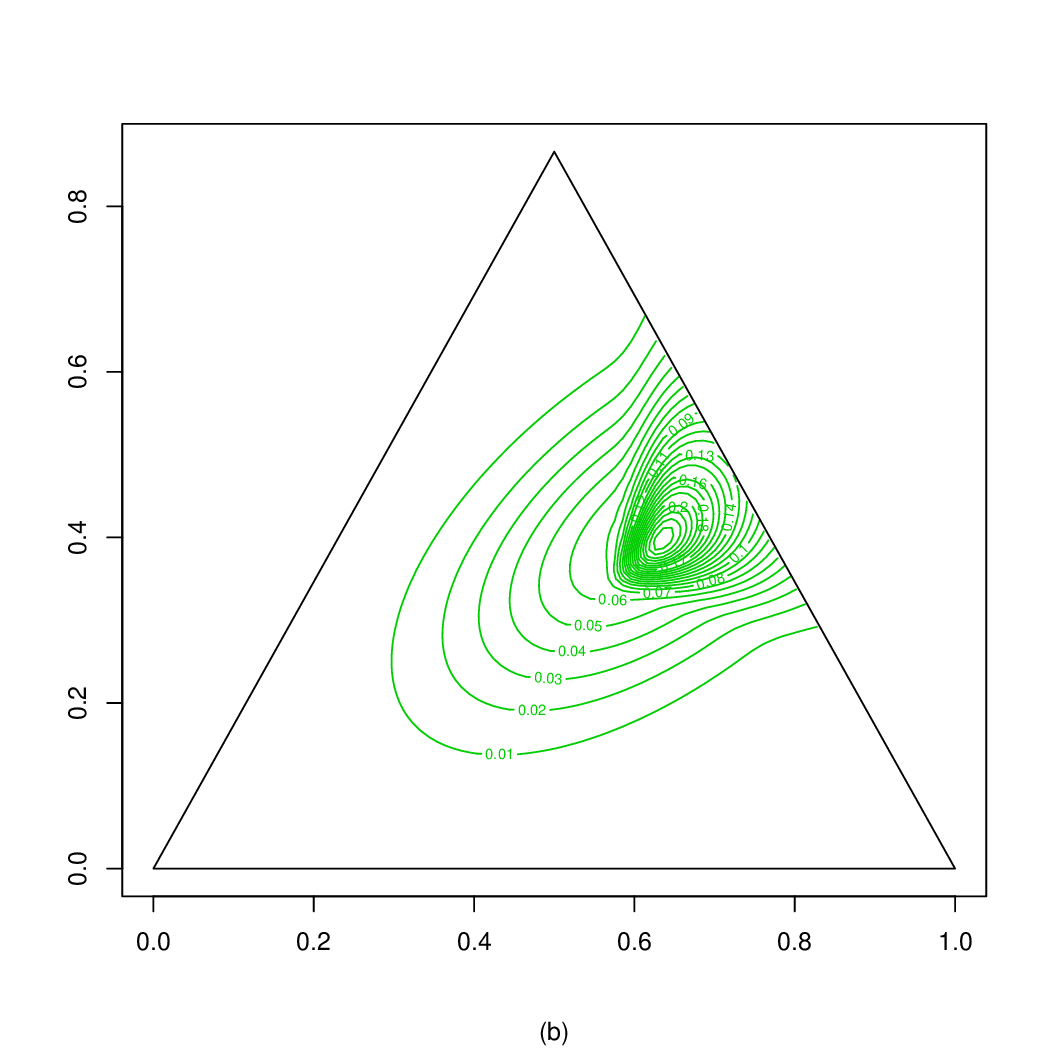} \\
(a)   &  (b)  \\
\includegraphics[scale=0.3,trim=0 0 0 0]{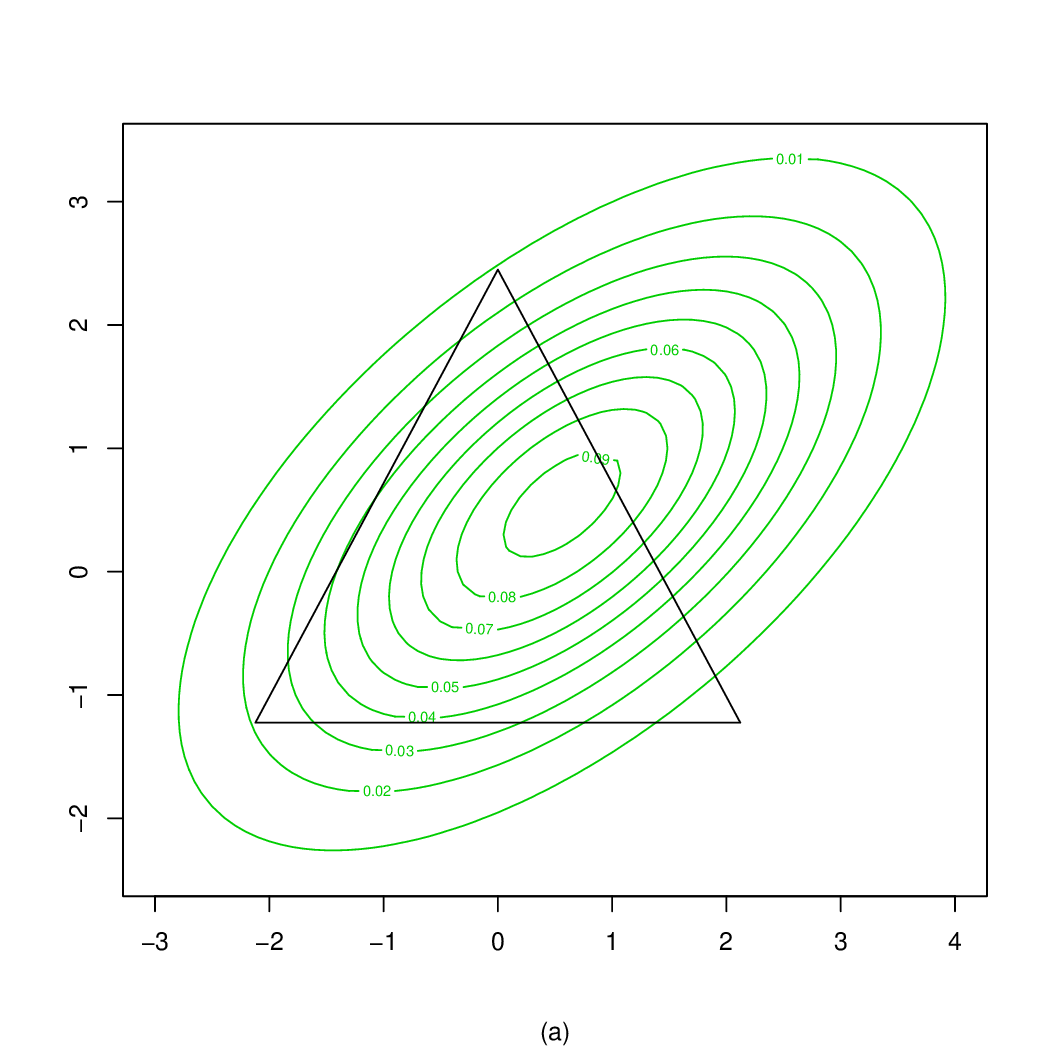} &
\includegraphics[scale=0.3,trim=0 0 0 0]{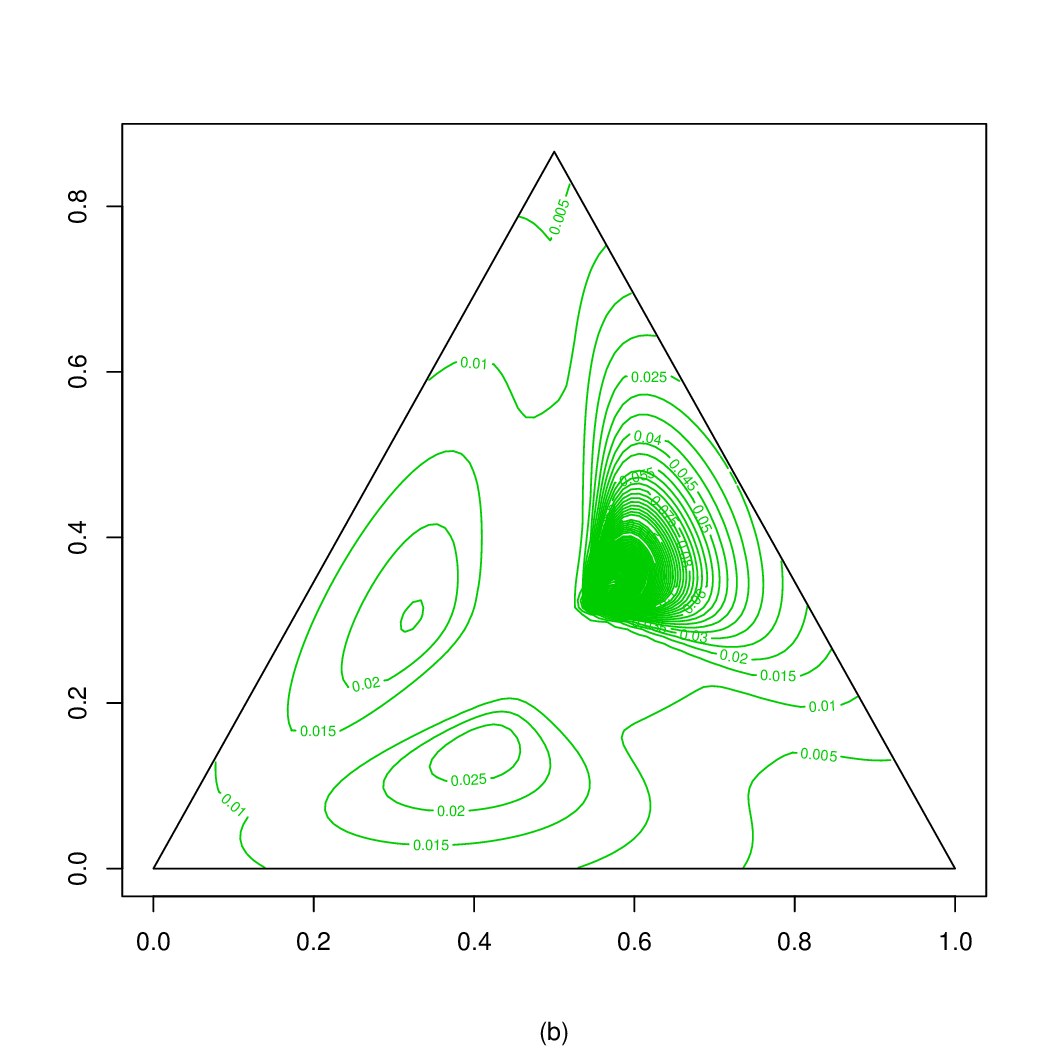} \\
(c)   &  (d)
\end{tabular}
\caption{All contour plots refer to a normal distribution with mean $\pmb{\mu}=\left(0.561, 0.547 \right)^T$. The covariance matrix of the first row is $\pmb{\Sigma}_1$ and of the second row is $\pmb{\Sigma}_2=5\pmb{\Sigma}_1$. The left plot is the normal distribution in real space and the triangle (the simplex after the $\alpha$-transformation in Equation (\ref{alpha}) with $\alpha=1$) is for illustration purposes. The left column shows the contour plots of the multivariate normal in $\mathbb{R}^2$ and the right column shows the contours of the $\alpha$-folded normal on the simplex.}
\label{contour1}
\end{figure}

For each combination of parameters we calculated the density of the normal and the folded normal for a grid of two-dimensional vectors and then plotted their contours. While for the unconstrained normal case, the density was simply calculated for all grid points, for the folded model, the transformation in Equation (\ref{trans}) was first applied (with $\alpha=1$) and then the density in Equation (\ref{foldednormal}) was calculated. The contour plots are presented in Figure \ref{contour1}.      

Figures \ref{contour1}(a) and \ref{contour1}(c) depict the contours of the multivariate normal distribution (defined in $\mathbb{R}^2$) while Figures \ref{contour1}(b) and \ref{contour1}(d) show the contours of the $\alpha$-folded normal (defined on the simplex). The first row corresponds to $\pmb{\Sigma}_1$ in Equation (\ref{Sigma}) while the second row is derived from  $\pmb{\Sigma}_2$. The triangles in the second column (Figures \ref{contour1}(b) and \ref{contour1}(d)) display the simplex while the corresponding triangles in Figures (a) and (c) were obtained through the $\alpha$-transformation in Equation (\ref{alpha}). 

What is perhaps most evident from the contour plots is that points falling outside the triangle in Figures \ref{contour1}(a) and \ref{contour1}(c) result in modes inside the simplex in Figures \ref{contour1}(b) and \ref{contour1}(d) respectively. When there is a high probability of being left outside of two or more sides of the triangle (or faces of a pyramid or hyper-pyramid in higher dimensions), as in Figures \ref{contour1}(c) and \ref{contour1}(d), the contours of the folding model will have a somewhat peculiar shape due to a multi-modal distribution arising on the simplex. The multi-modality depends upon the allocation of the probability left outside the simplex along the components. If only one side of the simplex has probability left outside (as in Figure \ref{contour1}(a)) then the resulting distribution will be unimodal (see Figure \ref{contour1}(b)). 

An estimate of the probability left outside each side of the triangle (or the simplex) may be obtained through simulation in a straightforward manner. To accomplish this for our example, we generated data from the multivariate normal distribution with the parameters previously specified and applied the inverse of the $\alpha$-transformation (that is, $g_0^\alpha(\mathbf{y})$) in Equation (\ref{trans}).  For the points left outside of the simplex, we simply calculated how many are outside of each edge of the triangle and divided by the total number of the simulated data points.

If we partition the missed probability into three parts, where each part refers to one of the three components, then we obtain the values $(0.008, 0.018, 0.124)$ for the case when $\pmb{\Sigma}=\pmb{\Sigma}_1$ (see Figures \ref{contour1}(a) and \ref{contour1}(b)).  In this case, most of the probability is left outside the third component and the total probability left outside of the simplex is therefore $0.15$. The total probability left outside of the simplex when $\pmb{\Sigma}=\pmb{\Sigma}_2$ (see Figures \ref{contour1}(c) and \ref{contour1}(d)) is $0.557$ and the allocation to the three components is different than in the previous example, namely ($0.141, 0.138, 0.278$). Since, in this case, all of the estimates are relatively high, multi-modality appears in Figure \ref{contour1}(d).

 \subsection{Maximum Likelihood Estimation} \label{mle}
The estimation of the parameters of the $\alpha$-folded model on $\mathbb{S}^2$ is not too complicated mainly because there are not too many parameters involved in the maximization of the log-likelihood. In particular, when $D=3$, there are $2$ parameters for the mean vector, $3$ for the covariance matrix as well as one for each of $\alpha$ and the probability $p$, and we can use the ``simplex'' algorithm of Nelder \& Mead (1965) to maximize the logarithm of Equation (\ref{foldednormal}), available via the command \textit{optim} in R (R Core Team, 2015). This algorithm is generally robust and is derivative free (Venables \& Ripley, 2002). However when moving to higher dimensions (roughly $D$ = 5 or larger), the maximization is not straightforward. 

For this reason, we use the EM algorithm (McLachlan \& Krishnan, 2007) to maximize the log-likelihood corresponding to Equation (\ref{foldednormal}). Let ${\bf X}$ denote the sample of $n$ compositional vectors in $\mathbb{S}^{D-1}$. Following Jung, Foskey \& Marron (2011) who applied the EM algorithm in the context of a univariate folded normal, we propose the algorithm below to estimate the unknown parameters ($\alpha$, $p$, $\pmb{\mu}_{\alpha}$ and $\pmb{\Sigma}_{\alpha}$) from a sample of compositional data. \ \\

\noindent
\underline{EM Algorithm}
\begin{enumerate}
\item[Step 1.] For a fixed value of $\alpha$, apply the $\alpha$-transformation without the Helmert sub-matrix multiplication (that is, Equation (\ref{alef})) to the compositional data ${\bf X}$ to obtain the matrix ${\bf W}_{\alpha}$.
\item[Step 2.] Calculate $w^*_i$ for each vector ${\bf w}^{\alpha}_i$, for $i=1,\ldots, n$.
\item[Step 3.] Left multiply each ${\bf w}^{\alpha}_i$ by the Helmert sub-matrix ${\bf H}$ to obtain ${\bf y}^{\alpha}_{0i}$. Then multiply each ${\bf y}^{\alpha}_{0i}$ by $\frac{1}{w^{*2}_i}$ to obtain ${\bf y}^{\alpha}_{1i}$ (see Equation (\ref{invtrans})). The ${\bf y}^{\alpha}_{0i}$ and ${\bf y}^{\alpha}_{1i}$ are the transformed compositional data onto $\mathbb{A}_{\alpha}^{D-1}$ and $\mathbb{R}^{D-1}\setminus \mathbb{A}_{\alpha}^{D-1}$, for $i=1,\ldots, n$ respectively.   Let $\mathbf{x}_{0i}^\alpha =g_0^\alpha(\mathbf{y}_{0i}^\alpha)$ and
$\mathbf{x}_{1i}^\alpha =g_1^\alpha(\mathbf{y}_{1i}^\alpha)$.

\item[Step 4.] Choose initial values for the estimates of the parameters, for example
\begin{eqnarray*}
 \hat{\pmb{\mu}}_{\alpha}^0=\frac{\sum_{i=1}^n{\bf y}^{\alpha}_{1i}}{n} \ \ \& \ \ \hat{\pmb{\Sigma}}_{\alpha}^0=\frac{1}{n}\sum_{i=1}^n\left({\bf y}^{\alpha}_{0i}-\hat{\pmb{\mu}}_{\alpha}^0\right)\left({\bf y}_{0i}-\hat{\pmb{\mu}}_{\alpha}^0\right)^T
\end{eqnarray*} 
and 
\begin{eqnarray*}
\hat{t}_i^0=\frac{f_{\mathbf{x}_0}(\mathbf{x}_{0i}^\alpha)}{f_{\mathbf{x}_0}(\mathbf{x}_{0i}^\alpha)+f_{\mathbf{x}_1}(\mathbf{x}_{1i}^\alpha)}  \ \
\& \ \ \hat{p}^0=\frac{\sum_{i=1}^n\hat{t}_i^0}{n}, 
\end{eqnarray*} 
where $\hat{t}_i$ is the estimated conditional expectation of the indicator function that indicates whether the $i$-th observation belongs to $f_{\mathbf{x}_0}$ or $f_{\mathbf{x}_1}$. 
\item[Step 5.] Update all the parameters each time, for $k \geq 1$
\begin{eqnarray*}
\hat{\pmb{\mu}}_{\alpha}^k &=& \frac{\sum_{i=1}^n\hat{t}_i^{k-1}{\bf y}^{\alpha}_{0i}+\sum_{i=1}^n\left(1-\hat{t}_i^{k-1}\right){\bf y}^{\alpha}_{1i}}{n} \\
\hat{\pmb{\Sigma}}_{\alpha}^k &=& \frac{1}{n} \left[ \sum_{i=1}^n\hat{t}_i^{k-1}\left({\bf y}^{\alpha}_{0i}-\hat{\pmb{\mu}}_{\alpha}^k\right)\left({\bf y}^{\alpha}_{0i}-\hat{\pmb{\mu}}_{\alpha}^k\right)^T \right.  \\
& & \left. + \sum_{i=1}^n\left(1-\hat{t}_i^{k-1}\right)\left({\bf y}^{\alpha}_{1i}-\hat{\pmb{\mu}}_{\alpha}^k\right)\left({\bf y}^{\alpha}_{1i}-\hat{\pmb{\mu}}_{\alpha}^k\right)^T\right] \\
\hat{t}_i^k &=& \frac{f_{\mathbf{x}0}(\mathbf{x}_{0i}^\alpha)}{f_{\mathbf{x}_0}(\mathbf{x}_{0i}^\alpha)+f_{\mathbf{x}_1}(\mathbf{x}_{1i}^\alpha)} \\
\text{and} \ \ \hat{p}^k &=& \frac{\sum_{i=1}^n\hat{t}_i^k}{n} 
\end{eqnarray*}
\item[Step 6.] 
Repeat Step 5 until the change between two successive log-likelihood 
\begin{eqnarray} \label{foldedmle}
\ell_{\alpha}=\sum_{i=1}^n\log\left[p f_{\mathbf{x}_0}(\mathbf{x}_{0i}^\alpha)+ 
\left(1-p\right) f_{\mathbf{x}_1}\left(\mathbf{x}_{1i}^\alpha\right)\right]
\end{eqnarray}
values is less than a tolerance value, where $f_{\mathbf{x}_0}$ and $f_{\mathbf{x}_1}$ are given in Equations (\ref{f0}) and (\ref{f1}) respectively.
\end{enumerate} 

\noindent The above described procedure should be repeated for a grid of values of $\alpha$, for example $(-1,-0.9,\ldots,0.9,1)$, and the value of $\alpha$ which maximizes the log-likelihood is chosen as its estimate. A more efficient search for the best $\alpha$ is via Brent's algorithm (Brent, 2013). When $\alpha=0$, the MLE estimates of the transformed data are obtained directly; no EM algorithm is necessary as all the probability is retained with the simplex. The fact that (\ref{alpha}) tends to (\ref{ilr}) as $\alpha \rightarrow 0$ ensures the continuity of the log-likelihood at $\alpha=0$. 

\subsection{Generating Data from the $\alpha$-Folded Multivariate Normal Distribution} \label{algorithm}

The algorithm below describes how to simulate a random vector from the $\alpha$-folded model in Equation (\ref{foldednormal}) when $\alpha \neq 0$.  The case when $\alpha =0$, is considered subsequently.\\
\begin{enumerate}
\item[Step 1.] Choose $\alpha,\ \pmb{\mu}$ and $\pmb{\Sigma}$, where $\alpha \neq 0$.
\item[Step 2.] Generate a $D-1$ by 1 vector $\mathbf{y}$ from a $N_{D-1}\left(\pmb{\mu}, \pmb{\Sigma}\right)$. 
\item[Step 3.] As per Equation (\ref{trans}), determine whether $\mathbf{y} \in \mathbb{A}^{D-1}$. To do this, compute $\mathbf{w} = \mathbf{H}^T\mathbf{y}$. If  $-\frac{1}{\alpha} \leq w_{i,\alpha} \leq \frac{D-1}{\alpha}$ for all components of $\mathbf{w}$ and $\sum_{i=1}^Dw_{i, \alpha}=0$, then $\mathbf{y} \in \mathbb{A}^{D-1}$ and let $\mathbf{x}=\mathbf{z}_\alpha^{-1}(\mathbf{y})$.  Otherwise, $\mathbf{y} \in \mathbb{R}^{D-1}\setminus \mathbb{A}^{D-1}$ and let $\mathbf{x} = {\bf w}_{\alpha}^{-1}\left( \frac{ \mathbf{H}^T\mathbf{y}}{ q^{*2}_\alpha(\mathbf{y}) }\right)$ where $q^{*}=\alpha\min \left\lbrace \mathbf{H}^T\mathbf{y} \right\rbrace$. 
\end{enumerate}

When $\alpha=0$, the following simplified algorithm is used:
\begin{enumerate}
\item[Step 1.] Choose $\pmb{\mu}$ and $\pmb{\Sigma}$.
\item[Step 2.] Generate a $D-1$ by 1 vector $\mathbf{y}$ from a $N_{D-1}\left(\pmb{\mu}, \pmb{\Sigma}\right)$. 
\item[Step 3.] Compute $\mathbf{w} = \mathbf{H}^{T}\mathbf{y}$.
\item[Step 4.] Using Equation (\ref{winv0}), compute
\[\mathbf{x} = \mathbf{w}_0^{-1}(\mathbf{w}).\]
\end{enumerate}


\subsection{Inference for $\alpha$}
In the previous two subsections, simplifications arose if $\alpha=0$ and it may, consequently, be worthwhile to test whether the simpler multivariate normal in Equation (\ref{ilr density}) (corresponding to $\alpha=0$) is appropriate for the data at hand.

Consider the hypothesis test: $H_0: \alpha=0$ versus $H_1: \alpha \neq 0$. While one option is to use a log-likelihood ratio test, depending on the alternative hypothesis (a. $\alpha\neq 0$ \& $p<1$, b. $\alpha\neq 0$ \& $p=1$, c. $\alpha=1$ \& $p<1$ or d.  $\alpha=1$, $p=1$), the degrees of freedom will vary. We have not encountered case d. so far in our data analyses but, in this case, would recommend using a Dirichlet model. In fact, if we generate data from a Dirichlet distribution, case d. is expected to arise from the MLE of Equation (\ref{foldednormal}). An alternative to the log likelihood ratio test is to use a parametric bootstrap such as the hypothesis testing procedure described below.

\begin{enumerate}
\item[Step 1.] For a given compositional data set, estimate the value of $\alpha$ obtained via the EM algorithm. This is the observed test statistic denoted by $\alpha_{obs}$.
\item[Step 2.] Apply the $\alpha$-transformation in Equation (\ref{alpha}) with $\alpha = \alpha_{obs}$ to the data. The data are now mapped onto set $\mathbb{A}_{\alpha}^{D-1}$ in Equation (\ref{Ad}). 
\item[Step 3.] Apply the inverse of the isometric transformation with $\alpha=0$ in Equation (\ref{ilr}) to the data in Step 2. to form a new sample of compositions acquired with $\alpha=0$. That is, the data has been transformed under the null hypothesis.
\item[Step 4.] Re-sample $B$ times from this new compositional data set and each time estimate the value of $\alpha_b$ for $b=1,\ldots,B$. 
\end{enumerate}
The $p-$value is then given by (Davison \& Hinkley, 1997)
\begin{eqnarray} \label{boot}
p-\text{value}=\frac{ \sum_{b=1}^B{\bf I}\lbrace b:\alpha_b\geq \alpha_{obs} \rbrace +1 }{B+1}, 
\end{eqnarray}
where $\bf I$ is the indicator function.

One might argue that the value of $\alpha$ itself is not a pivotal statistic, in fact it is not even standardized, so a second bootstrap should be performed to obtain the standard error of the estimate for each bootstrap sample. In order to avoid this extra computational burden, the parametric bootstrap hypothesis testing could alternatively be carried out using the log-likelihood ratio test statistic in Steps 1 and 4 above. 

Inference for $\alpha$ could also be achieved via the construction of bootstrap confidence intervals. For this approach, simply re-sample the observations (compositional vectors) from the compositional data set and find the value of $\alpha$ for which the log-likelihood derived from Equation (\ref{foldednormal}) is maximized for each bootstrap sample. By repeating this procedure many times, we can empirically estimate the distribution of $\hat{\alpha}$, including its standard error. A variety of confidence intervals may be formed based on this distribution (see Davison \& Hinkley, 1997 and the R package \textit{boot}). The percentile method, for example, simply uses the $2.5\%$ lower and upper quantiles of the bootstrap distribution as confidence limits.

A less computationally intensive approach to obtain confidence intervals is based upon the second derivative of the profile log-likelihood of $\alpha$, that is, the observed Fisher's information measure. Assuming asymptotic normality of the estimator, the inverse of the observed information serves as an estimate of the standard error of the maximum likelihood estimator (Cox \& Hinkley, 1979). 

\section{Data Analysis Examples}  \label{data}

\subsection{Example 1: Sharp's Data Set I}  \label{SharpI}
Our first example makes use of Sharp's (2006) first $25 \times 3$ artificial data set, (termed "3a" by Sharp), made up from Aitchison's Hongite data (Aitchison, 2003). We chose to analyze Sharp's artificial data (Sharp, 2006) because they are curved data and according to Aitchison (2003) the logistic normal in Equation (\ref{ilr density}) should produce a very good fit for curved data. Clearly a Dirichlet distribution would fail to capture the variability of such data and we would not expect a value of $\alpha=1$ to do better. Sharp (2006) showed that the normalized geometric mean of the components (assuming a logistic normal distribution) fails to lie within the corpus of the data, whereas the spatial graph median does. 

\begin{figure}[!ht]
\centering
\begin{tabular}{cc}
\includegraphics[scale=0.47,trim=0 30 0 20]{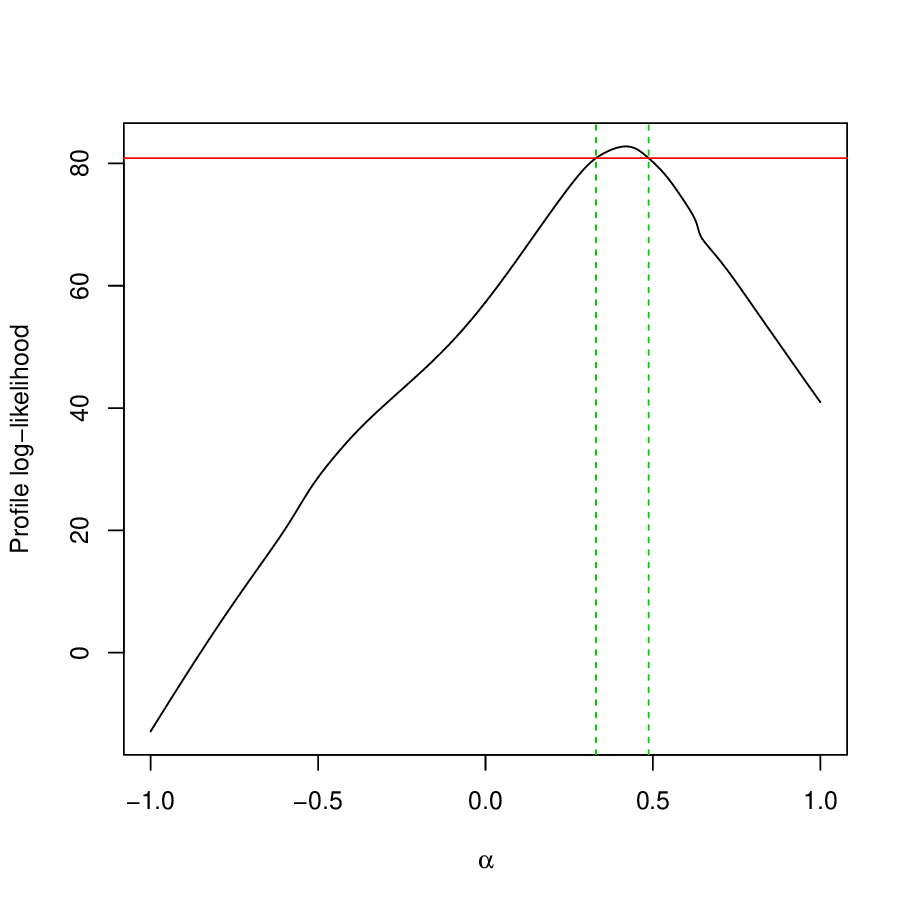} &
\includegraphics[scale=0.44,trim=30 50 0 20]{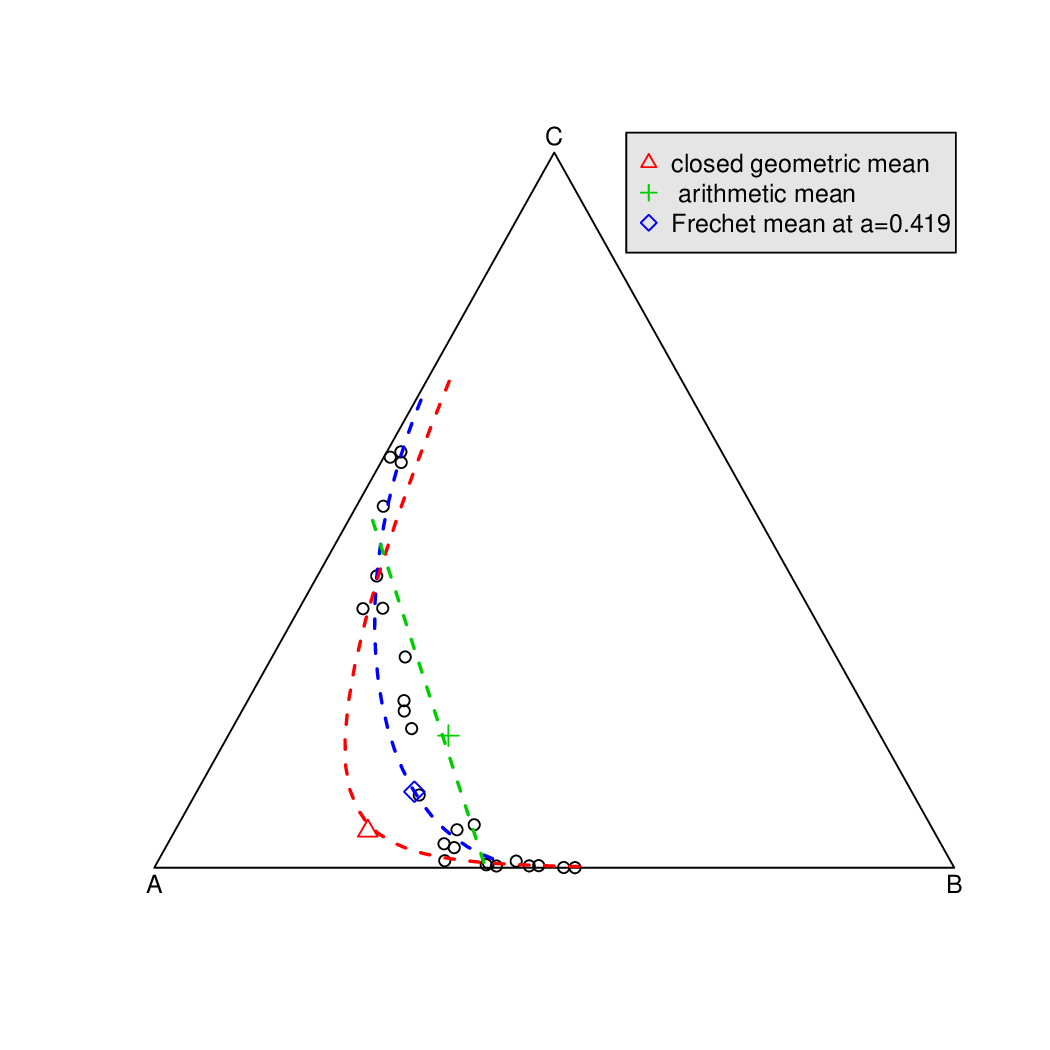} \\
\footnotesize{(a)}   &  \footnotesize{(b)}  
\end{tabular}
\caption{\textbf{Sharp's Data Set I} (a) Profile log-likelihood of $\alpha$. The red and green lines indicate the $95\%$ confidence interval of the true value of the parameter. (b) Ternary plot of the data along with three $\alpha$-means evaluated at $\alpha=0$ (geometric mean normalised to sum to 1), $\alpha=0.419$ and $\alpha=1$ (arithmetic mean). The lines correspond to the scores of the first principal component for each value of $\alpha$. }
\label{fold1}
\end{figure}

Figure \ref{fold1}(a) shows the profile log-likelihood of $\alpha$ and the maximum of the log-likelihood which occurs at $\alpha=0.419$. The log-likelihood values at $\alpha=0.419$ and $\alpha=0$ are equal to $82.780$ and $57.316$ respectively. The log-likelihood ratio test based on a $\chi^2$ distribution with $2$ degrees of freedom clearly rejects the logistic normal on the simplex (that is, that $\alpha=0$ is the optimal transformation) and this conclusion is in line with the confidence interval limits. For this example, $p$ was estimated to be approximately 0.95 so the probability of a point needing to be folded into the simplex was small (about 0.05) and the optimal value of $\alpha$ without folding, as proposed by Tsagris et al. (2011), was equal to $0.428$.

Figure \ref{fold1}(b) shows the ternary diagram of the data.
In order to obtain the displayed $\alpha$-mean, termed the Fr{\'e}chet mean by (Tsagris, Preston \& Wood, 2011), the EM algorithm estimate of $\pmb{\mu}_{0.419}$, say $\bar{{\bf y}}_0.419$ is transformed inside the simplex using Equation (\ref{trans}). When $\alpha=0$, the closed geometric mean can be similarly obtained through the inverse of the ilr transformation applied to $\bar{{\bf y}}_0$.  The closed geometric mean, arithmetic mean and the Fr{\'e}chet $\alpha$-mean in $\mathbb{S}^2$, are respectively 

\begin{eqnarray*}
\hat{\pmb{\mu}}_0 &=& \left(0.707, 0.241, 0.051\right) \ \ \left(\text{Normalized geometric mean} \right) \\
\hat{\pmb{\mu}}_1 &=& \left(0.540, 0.275, 0.185\right) \ \ \left(\text{Simple arithmetic mean} \right) \\
\hat{\pmb{\mu}}_{0.419} &=& \left( 0.622, 0.272, 0.106\right) \ \  \left(\text{Fr{\'e}chet mean using the proposed folded model} \right)
\end{eqnarray*}

Note that the Fr{\'e}chet $\alpha$-mean with $\alpha = 0.428$, corresponding to Tsagris et al. (2011), is $\hat{\pmb{\mu}}_{0.428} =\left(0.619, 0.271, 0.110 \right)$ which, as expected, is very similar to $\hat{\pmb{\mu}}_{0.419}$ for this example.  We can clearly see that both the simple and the closed geometric mean fail to lie within the main bulk of the data. However, the Fr{\'e}chet $\alpha$-mean calculated at $\alpha=0.419$ achieves this goal.

\begin{figure}[!ht]
\centering
\begin{tabular}{cc}
\includegraphics[scale=0.45,trim=0 60 0 20]{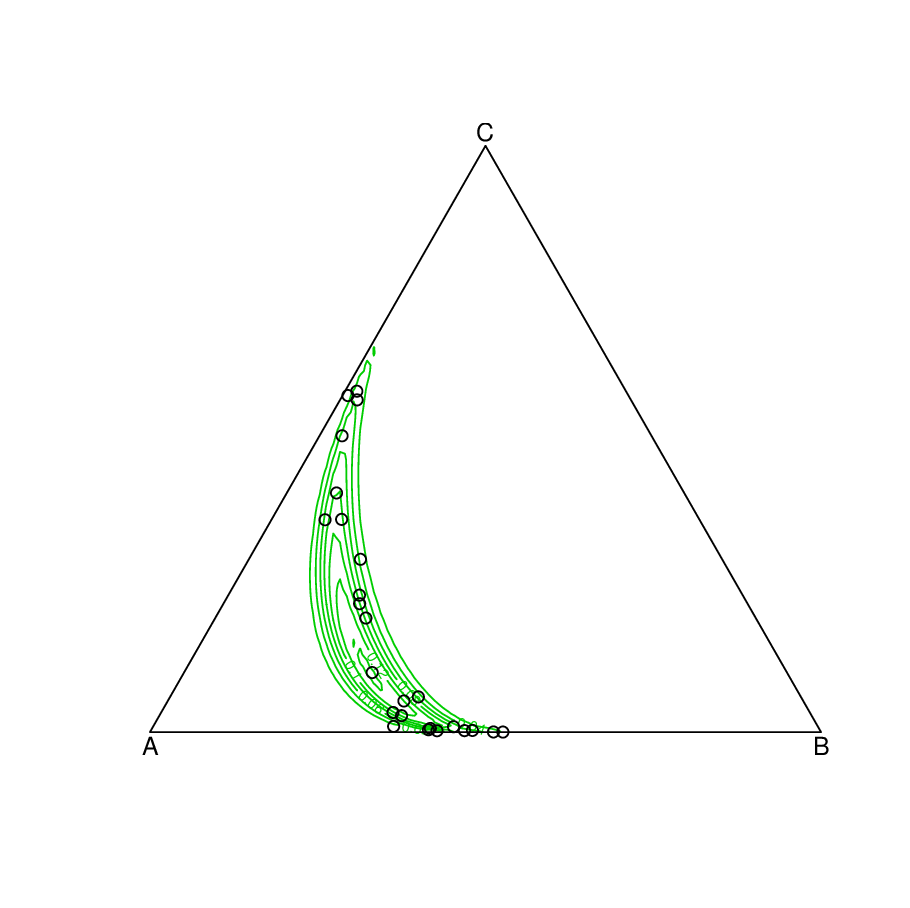} &
\includegraphics[scale=0.45,trim=0 60 0 20]{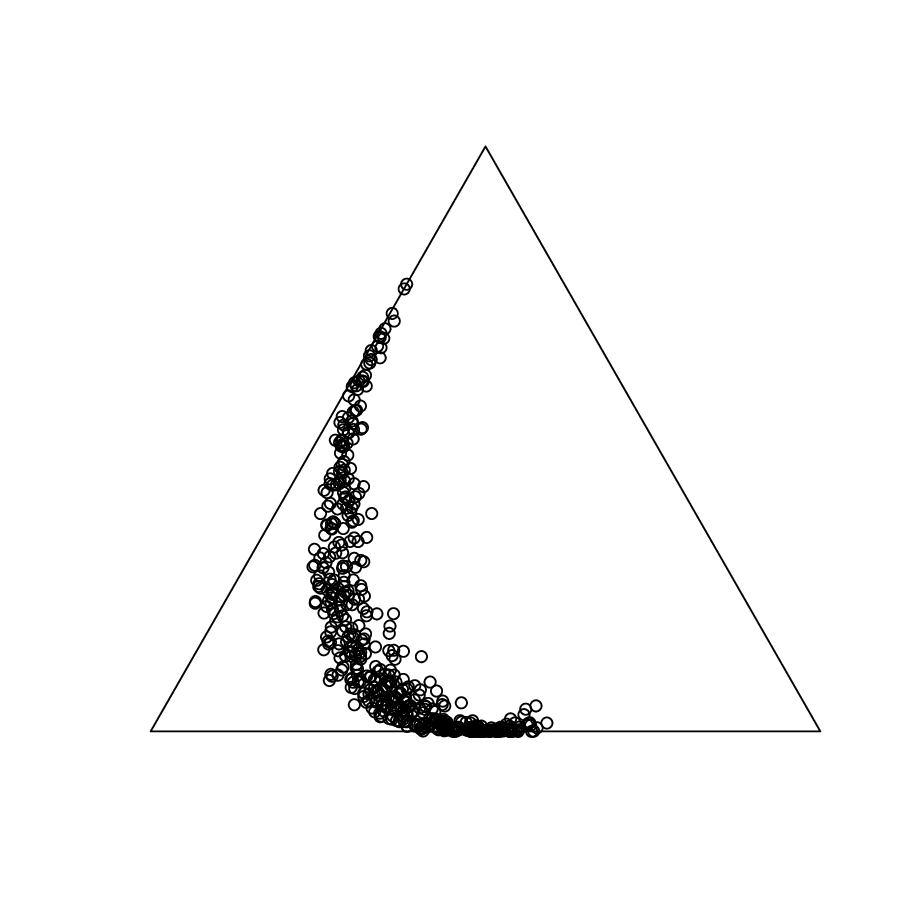} \\
\footnotesize{(a)}   &  \footnotesize{(b)}    \\
\includegraphics[scale=0.45,trim=0 60 0 20]{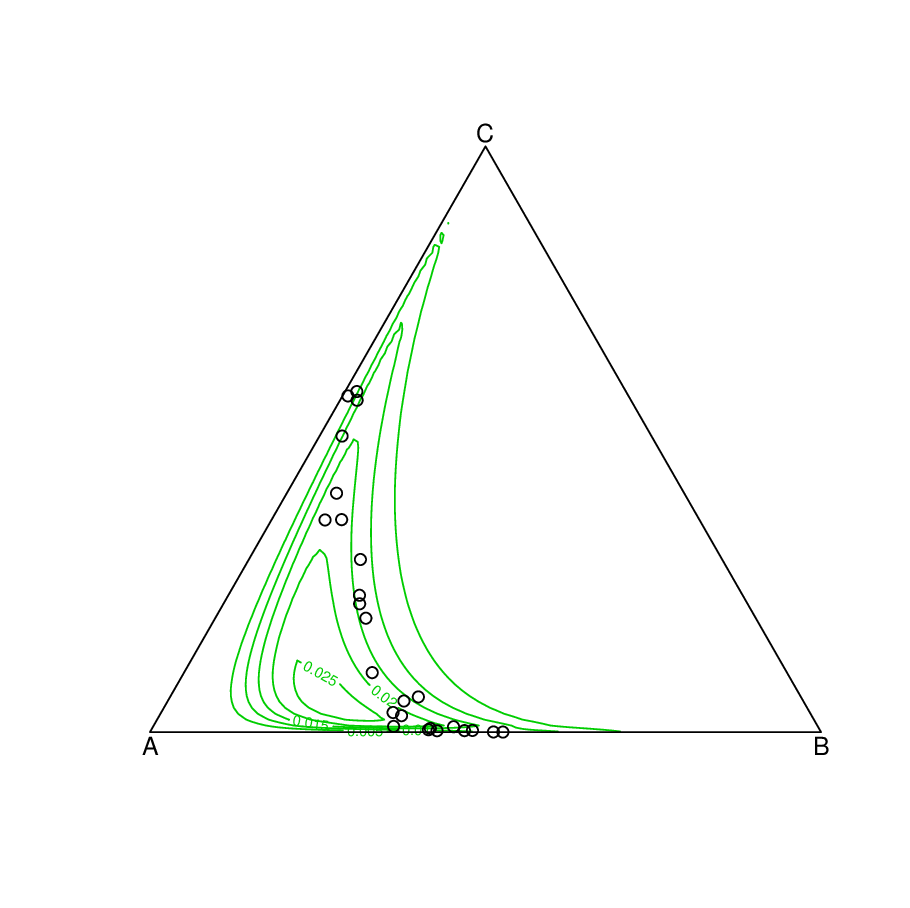} &
\includegraphics[scale=0.45,trim=0 60 0 20]{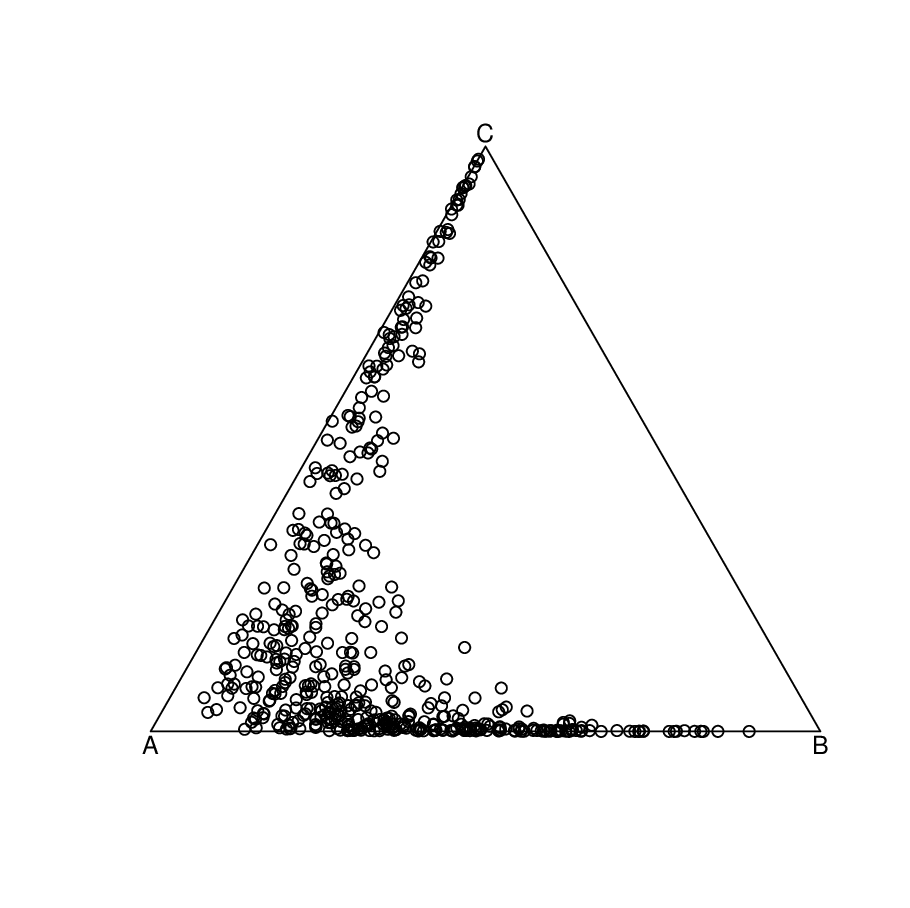} \\
\footnotesize{(c)}   &  \footnotesize{(d)}  
\end{tabular}
\caption{\textbf{Sharp's Data Set I} Contours of (a) the $\alpha$-folded model with $\alpha=0.419$ and (c) of the multivariate normal distribution applied to the $\alpha$-transformed data with $\alpha=0$. Plots (b) and (d) are $500$ simulated data from the $\alpha$-folded model with $\alpha=0.419$ and from the multivariate normal distribution applied to the $\alpha$-transformed data with $\alpha=0$ respectively.}
\label{cont1}
\end{figure}

The contour plots of the $\alpha$-folded model with $\alpha=0.419$ and $\alpha=0$ appear in Figures \ref{cont1}(a) and \ref{cont1}(c) respectively. We generated $500$ observations from each model and these are plotted in Figures \ref{cont1}(b) and \ref{cont1}(d). When $\alpha=0.419$ the simulated data look more like the observed data, in contrast to the simulated data with $\alpha=0$. 

The first principal component of the $\alpha$-transformed data for each value of $\alpha$ is also plotted. Principal component analysis for compositional data has been described by Aitchison (1983). The centered log-ratio transformation in Equation (\ref{clr}) is applied to the compositional data and standard eigen analysis is applied to the covariance matrix (which has at least one zero eigenvalue). If $\alpha \neq 0$, we suggest an analogous approach in which the estimated covariance matrix is mapped to $\mathbb{Q}_{\alpha}^{D-1}$ space (Equation (\ref{Qad})), using the Helmert sub-matrix as follows
\begin{eqnarray*}
\hat{\pmb{\Sigma}}_{\alpha}^*={\bf H}^T\hat{\pmb{\Sigma}}_{\alpha}{\bf H}.
\end{eqnarray*}
If $\alpha=0$, the covariance is mapped to $\mathbb{Q}_0^{D-1}$ (Equation (\ref{Qd})). Step 3 of the algorithm presented in Subsection \ref{algorithm} is used to back-transform the principal components onto the simplex. 

\subsection{Example 2: Sharp's Data Set II} \label{SharpII}
\begin{figure}[!ht]
\centering
\begin{tabular}{ccc}
\includegraphics[scale=0.4,trim=30 20 0 20]{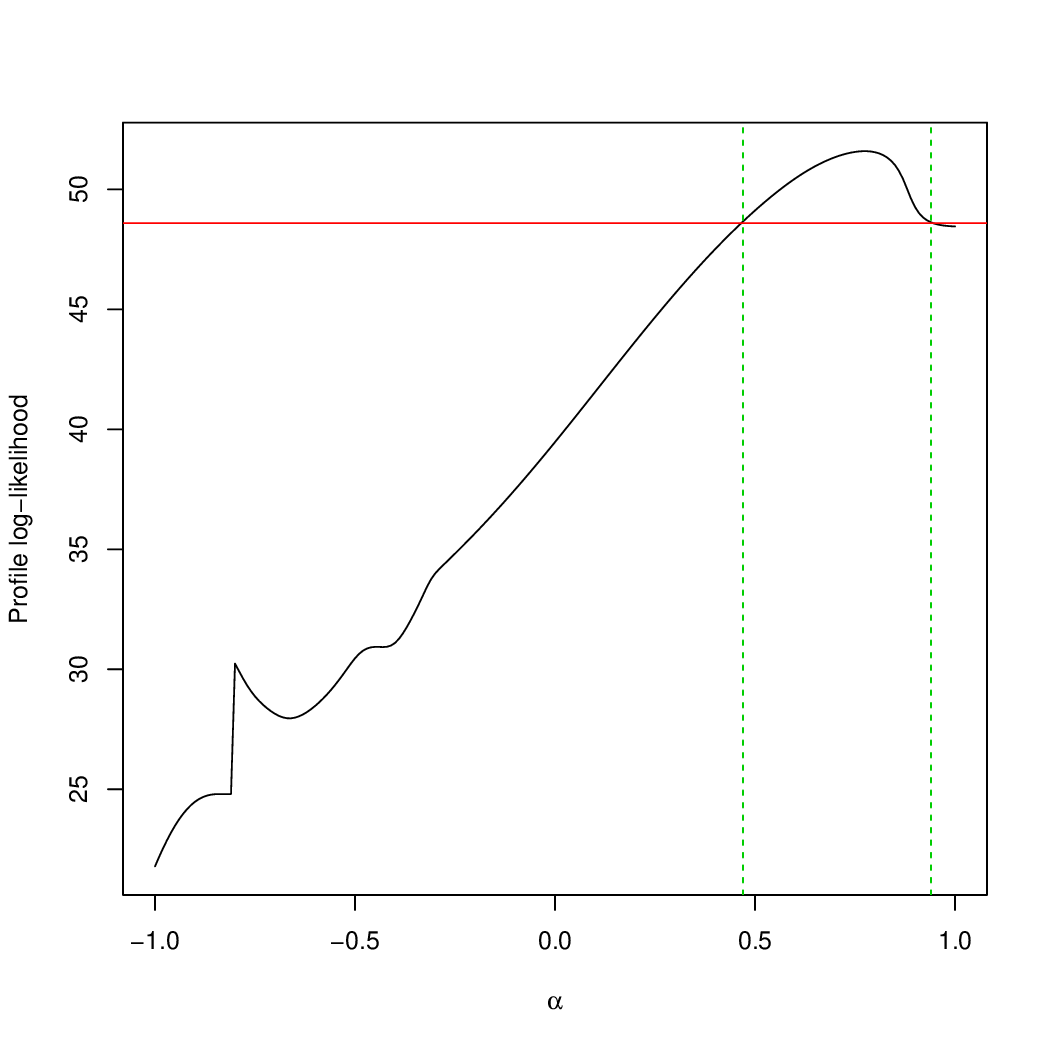} & 
\includegraphics[scale=0.45,trim=30 50 0 0]{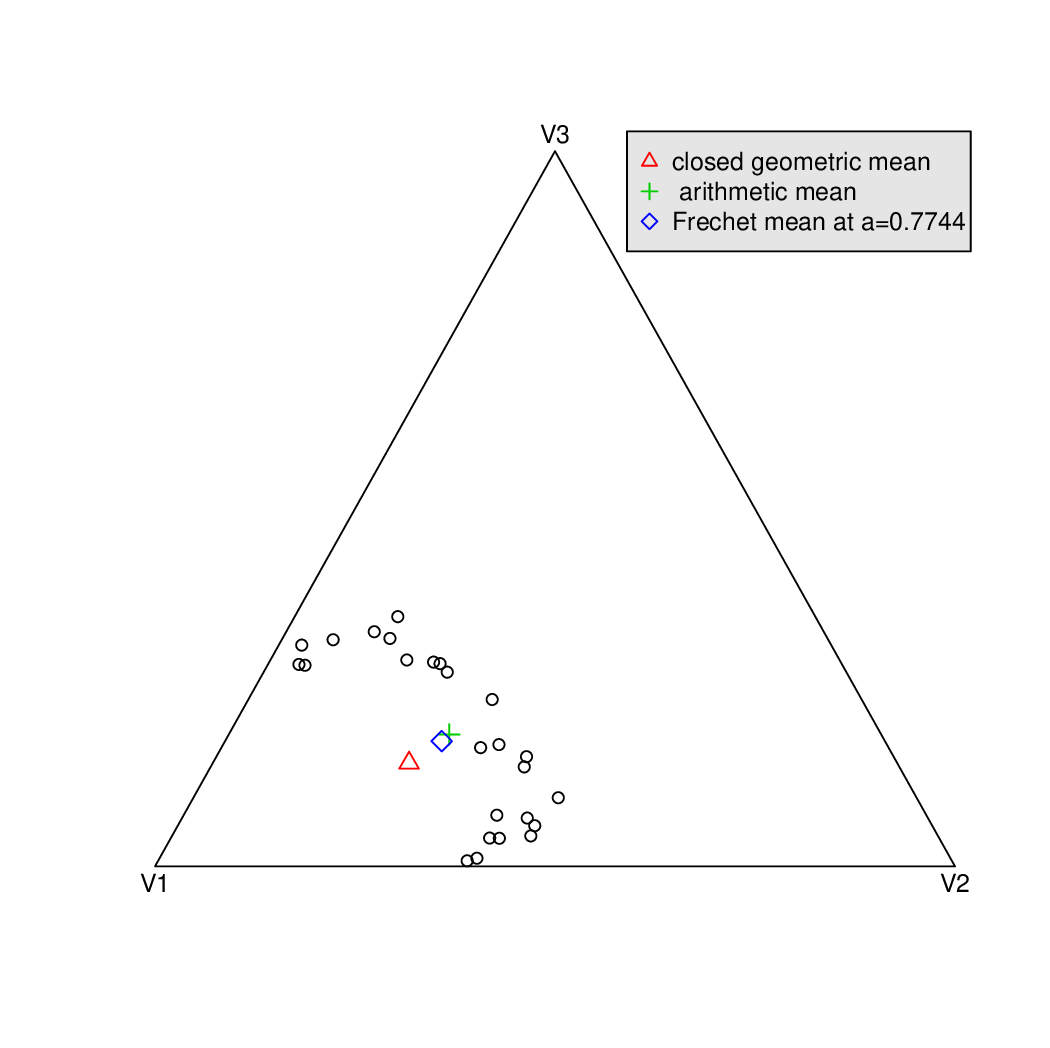} \\
\footnotesize{(a)}   &  \footnotesize{(b)}   \\
\includegraphics[scale=0.45,trim=40 60 0 20]{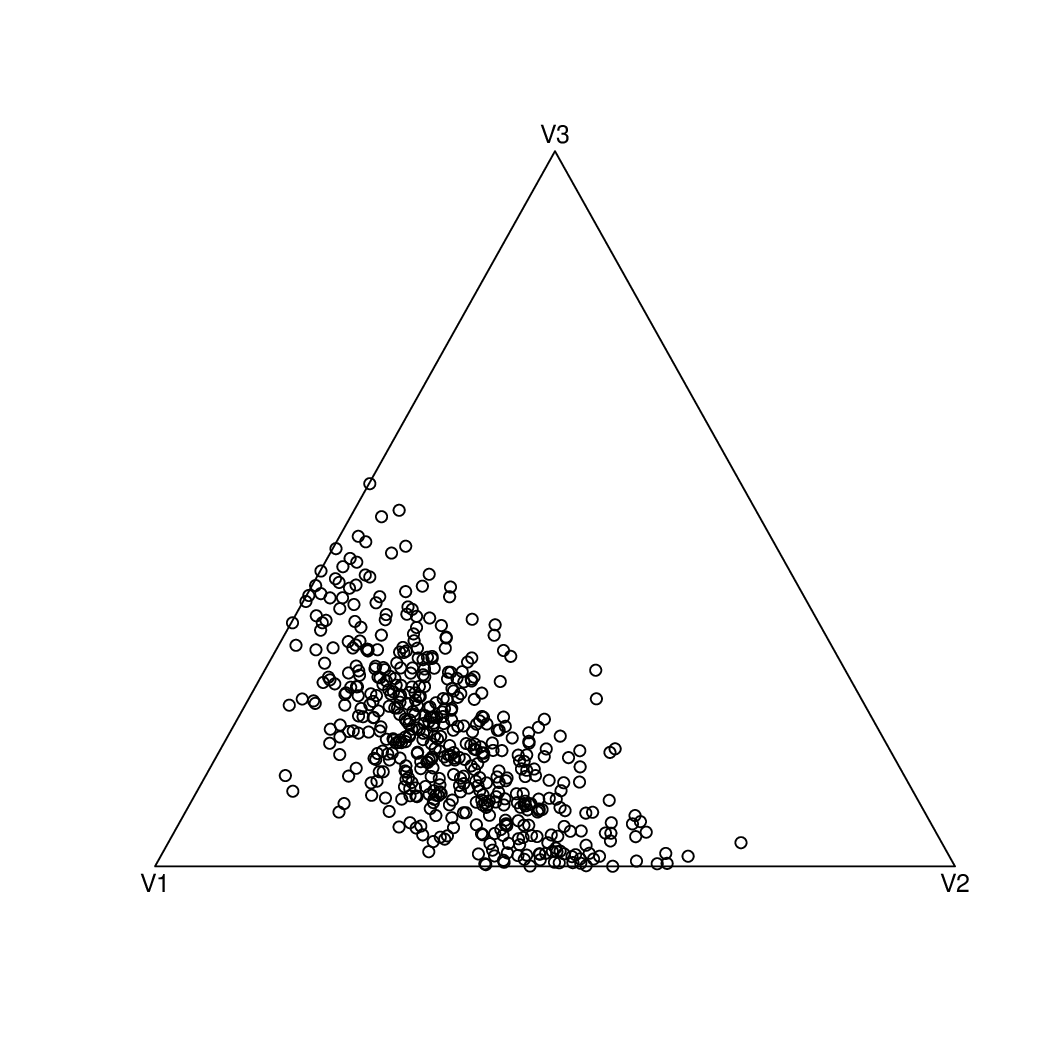} & 
\includegraphics[scale=0.45,trim=40 60 0 20]{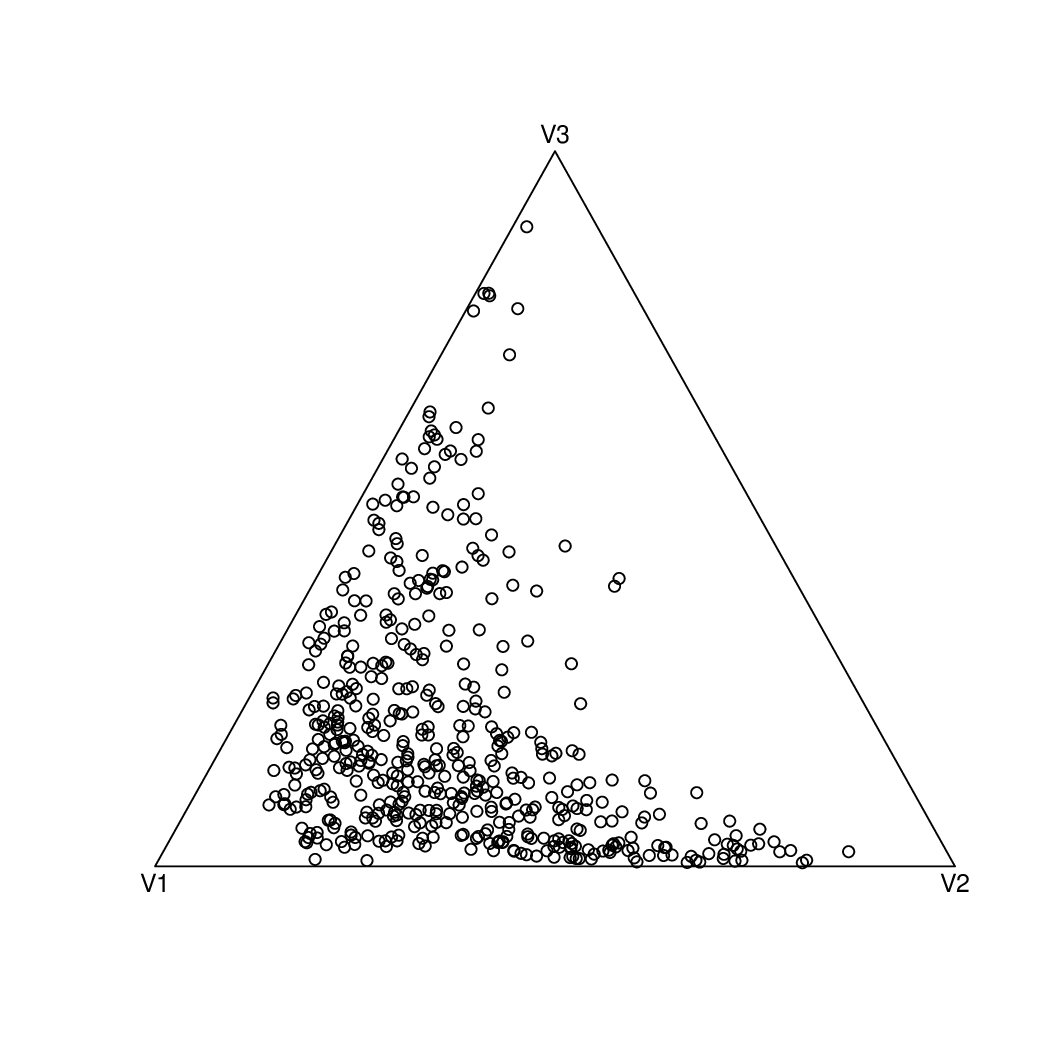}  \\
\footnotesize{(c)}  &  \footnotesize{(d)}  \\
\end{tabular}
\caption{\textbf{Sharp's Data Set II} (a) Profile log-likelihood of $\alpha$. The red and green lines indicate the $95\%$ confidence interval of the true value of the parameter. (b) Ternary plot of the data along with three Fr{\'e}chet-means evaluated at $\alpha=0$ (geometric mean normalised to sum to 1), $\alpha=0.774$ and $\alpha=1$ (arithmetic mean). Simulated data from the folded model with (c) $\alpha=0.774$ and (d) $\alpha=0$.}
\label{4figs}
\end{figure}

We analyzed a second $25 \times 3$ artificial data set taken from Aitchison's Hongite data (Aitchison, 2003), termed ``3c'' by Sharp (2006). As with the previous case, the data are also curved. Figure \ref{4figs} presents the relevant graphical information about the estimation of the fitted $\alpha$-folded model, namely the profile log-likelihood of $\alpha$, the ternary diagram of the data, and simulated data from the fitted $\alpha$-folded model as well as from the multivariate normal distribution applied to the $\alpha$-transformed data with $\alpha=0$.

This is an example where the $\alpha$-folded model has failed to capture the structure of the data. However, its competing model (the multivariate normal distribution applied to the isometric log-ratio transformed data) is clearly an even less attractive model for these data. 

In terms of the potential benefit of folding for this example, the estimated probability left outside of the simplex was only $0.02$ ($\hat{p}=0.98)$ and the optimal $\alpha$ using the model in Tsagris et al. (2011) without folding  was equal to 1, compared to $0.774$ with folding. The Fr{\'e}chet mean evaluated at $\alpha=0.774$ was equal to $(0.553, 0.270, 0.177)$ while the Fr{\'e}chet mean evaluated at $\alpha=1$ was similar and equal to $(0.540, 0.276, 0.184)$.

\subsection{Example 3: Coffee Aroma Data}
This data set contains more components (is of a higher dimensionality) than Examples 1 and 2. Thirty commercially available roasted coffee samples of different origins (Arabica, Robusta and various blends of them), processed by different technologies,  were analyzed by Korho{\v{n}}ov{\'a} et al. (2009).  In this example, we consider the six compounds (or compositional parts) selected by Korho{\v{n}}ov{\'a} et al., (2009). The estimated optimal value of $\alpha$ for this data set based on the folded model was equal to $0.908$, suggesting an improved fit over the logistic normal distribution. The estimated probability of an observation being left outside the simplex was relatively small (0.0523) and, consequently, the other parameter estimates were similar between those produced by the model introduced here with folding and the model in Tsagris et al. (2011) without.  

\subsection{Example 4: Labor Force Data}
The fourth data set comes from economics and also contains 6 components, as in Example 3. In particular, it contains information on the labour force by status in employment (6 variables) for 124 countries. The data set is accessible via the R package \textit{robCompositions} \cite{templ2011}. 

The analysis of this data set suggests that, as in the previous examples, an $\alpha$ value other than zero may provide an improved fit.  The optimal value of $\alpha$ using the model without folding in Tsagris et al. (2011) was equal to $0.328$, whereas its value using the proposed folded model was equal to $0.516$.  Furthermore, there appears to be a need for the folding transformation as the estimated probability of a point being left outside the simplex was large and equal to $0.756$.  While the two Fr{\'e}chet mean vectors were roughly equal ( $\pmb{\mu}_{0.328}=(0.350, 0.366, 0.010, 0.024, 0.122, 0.127)$ and $\pmb{\mu}_{0.516}=(0.348, 0.357, 0.011, 0.025, 0.129, 0.130)$), this is not surprising since as the number of components (or dimensionality) grows, the volume of the simplex becomes smaller and the mean vectors will be close for a wide range of values of $\alpha$. The difference, however, between the two models can be observed in the estimated covariance matrices which are given below.

\begin{eqnarray*}
\begin{array}{cc}
\hat{\pmb{\Sigma}}_{0.328} = \left( \begin{array}{ccccc}
0.071 & 0.086 & 0.058 & 0.249 & 0.150 \\
0.086 & 0.538 & 0.340 & 0.557 & 0.356 \\
0.058 & 0.340 & 0.255 & 0.367 & 0.233 \\
0.249 & 0.557 & 0.367 & 1.854 & 1.242 \\ 
0.150 & 0.356 & 0.233 & 1.242 & 0.906 \\
\end{array}\right)
&
\hat{\pmb{\Sigma}}_{0.516} = \left( \begin{array}{ccccc}
0.101 & 0.355 & 0.219 & 0.402 & 0.219 \\
0.355 & 2.627 & 1.574 & 2.368 & 1.499 \\
0.219 & 1.574 & 0.987 & 1.493 & 0.940 \\
0.402 & 2.368 & 1.493 & 3.351 & 2.171 \\
0.219 & 1.499 & 0.940 & 2.171 & 1.522 \\
\end{array}\right)
\end{array}
\end{eqnarray*}

The elements of $\hat{\pmb{\Sigma}}_{0.516}$ (with folding) are larger than those of $\hat{\pmb{\Sigma}}_{0.328}$ (without folding), ranging from 1.4 times up to 4.8 times larger and the generalised variance of $\hat{\pmb{\Sigma}}_{0.516}$ is $7$ times that of the first covariance matrix. These differences are the due to $\hat{p}$ being relatively small in this example.

\section{Simulation Studies}  \label{sim}

\subsection{Estimation of $\pmb{\mu}_\alpha$, $\pmb{\Sigma}_\alpha$ and $p$}
\label{sim1}
In this simulation study, $\alpha$ was fixed and we examined the accuracy of the EM algorithm in terms of estimating $\pmb{\mu}_\alpha$, $\pmb{\Sigma}_\alpha$ and $p$ for increasing sample sizes. Specifically, two values of $\alpha$ were chosen, namely $-0.5$ and $0.5$, and 4-dimensional data ($D=5$) from a multivariate normal distribution with two different mean vectors and a variety of different covariance parameters were generated. In particular, we used mean vectors
\begin{eqnarray*}
\pmb{\mu}_{-0.5}=\left(1.715, 0.914, 0.115, 0.167\right) \ \ \text{and} \ \ 
\pmb{\mu}_{0.5}=\left(-0.566, -0.979, -0.648, -0.651\right),  
\end{eqnarray*} 
and covariance matrices 
\begin{eqnarray} \label{covk}
\pmb{\Sigma}=\kappa\left( \begin{array}{cccc}
0.149  & -0.458 & 0.002  & -0.005 \\
-0.458 & 1.523  & 0.000  & 0.007  \\
0.002  & 0.000  & 0.037  & -0.047 \\
-0.005 & 0.007  & -0.047 &  0.061 \end{array}\right)
\end{eqnarray}
where $\kappa=0.5,1,2,3,5,7,10$. Note that the value of $\kappa$ changes the probability that a point is left outside of the simplex. The ``true'' values of $p$ for each value of $\kappa$ were computed through Monte-Carlo simulations by generating many random vectors from a multivariate normal distribution with the parameters above and computing the proportion of vectors that belong to $\mathbb{A}^{D-1}$. We will refer to the probability that a vector is outside of  $\mathbb{A}^{D-1}$ (that is, $1-p$) as the probability left outside the simplex since these vectors need to be folded into the simplex. This estimated probability, for each $\alpha$ and $\kappa$ combination, is presented in Table \ref{table1}.

For each combination of $\alpha$, $\pmb{\mu}$ and $\pmb{\Sigma}$, seven sample sizes, namely $n=(50,100,200,300,500,750, 1000)$, were considered. Results are based on $1000$ simulated data sets (for each $n$) and, for each simulated sample, estimates of $p$, $\pmb{\mu}$ and $\pmb{\Sigma}$ were calculated.  All computations took place in R 3.2.3 R (R Core Team, 2015) using a desktop computer with Intel Core i5 at 3.5 GHz processor and 32GB RAM. For various measures of distance (as described below), the mean distance between the 1000 estimates and the true parameters was calculated. 

For the probability left outside of the simplex (that is, $1-p$), the absolute difference between the estimated probability and the true probability was computed. For the mean vector, the Euclidean distance was calculated to measure the discrepancy between the estimated vector and true vector whereas for the covariance matrix, the following metric (F{\"o}rstner \& Moonen, 2003) was calculated
\begin{eqnarray} \label{covdist}
d\left(\hat{\pmb{\Sigma}},\pmb{\Sigma}\right)=\sqrt{\sum_{i=1}^{D-1}\left[\log{\Lambda_i\left(\hat{\pmb{\Sigma}}\pmb{\Sigma}^{-1}\right)} \right]^2},
\end{eqnarray}
where $\Lambda_i\left({\bf A}\right)$ denotes the $i$-th eigenvalue of the matrix $\bf A$.

\begin{figure}[!ht]
\centering
\begin{tabular}{ccc}
\multicolumn{3}{c}{Simulation study when $\alpha=-0.5$} \\ 
\includegraphics[scale=0.35,trim=50 0 0 30]{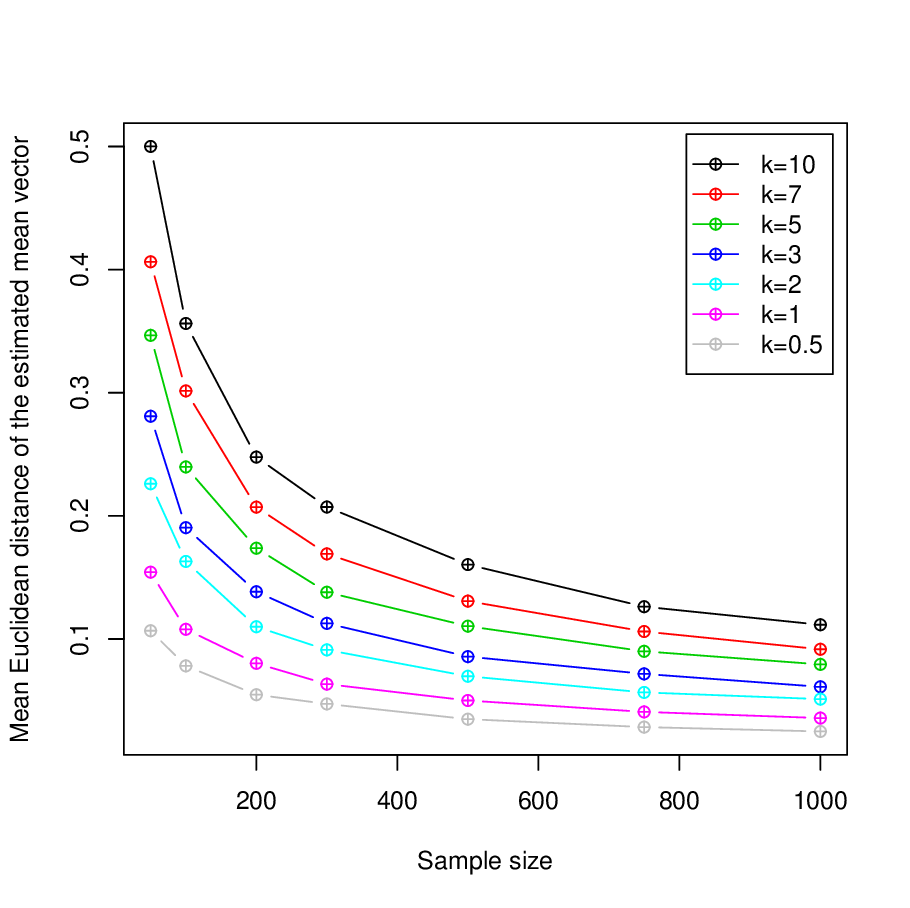} &
\includegraphics[scale=0.35,trim=50 0 0 30]{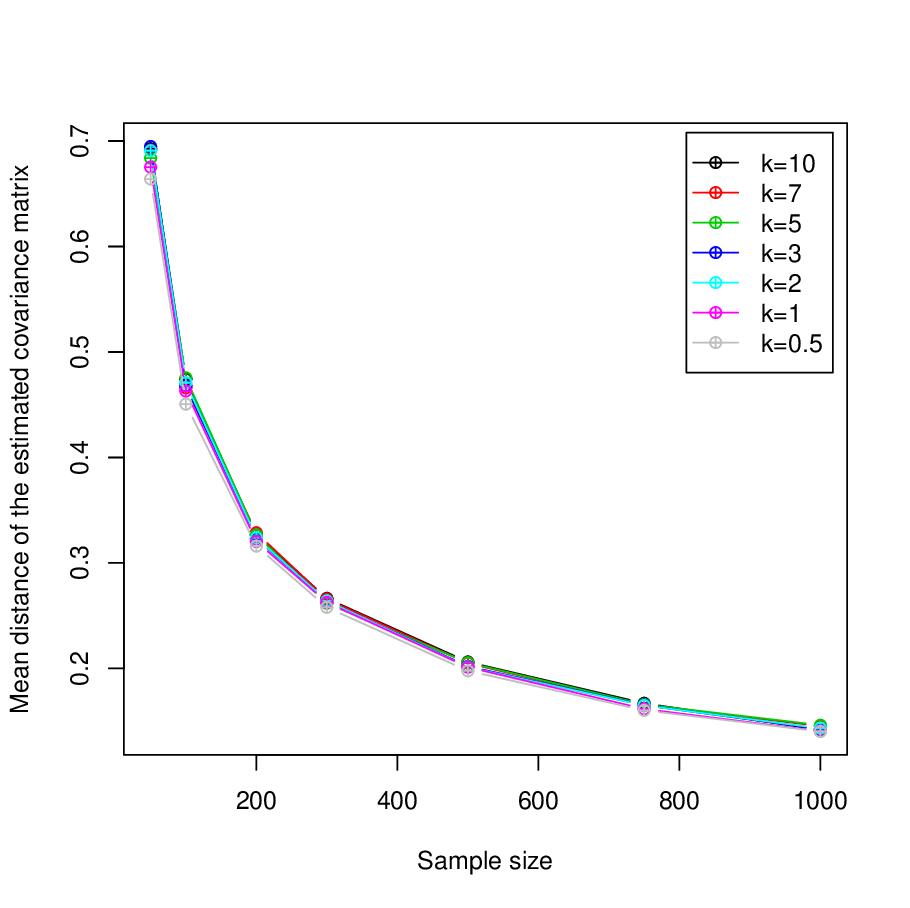} &
\includegraphics[scale=0.35,trim=50 0 0 30]{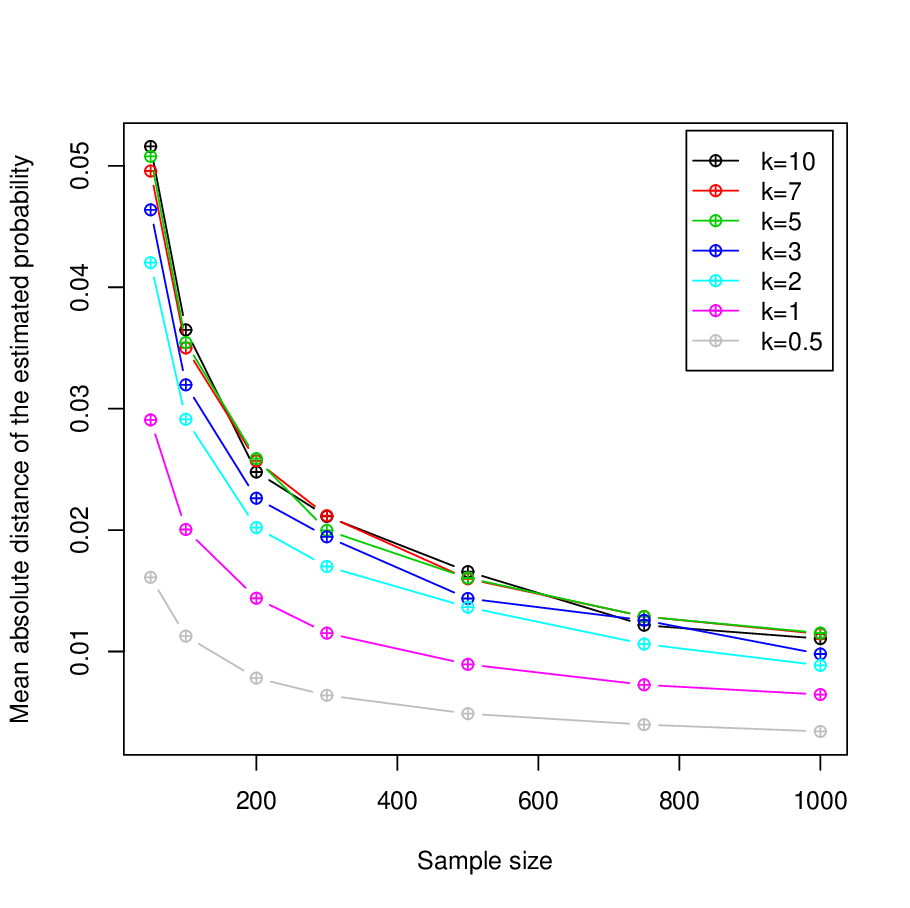} \\
\multicolumn{3}{c}{Simulation study when $\alpha=0.5$} \\ 
\includegraphics[scale=0.35,trim=50 0 0 30]{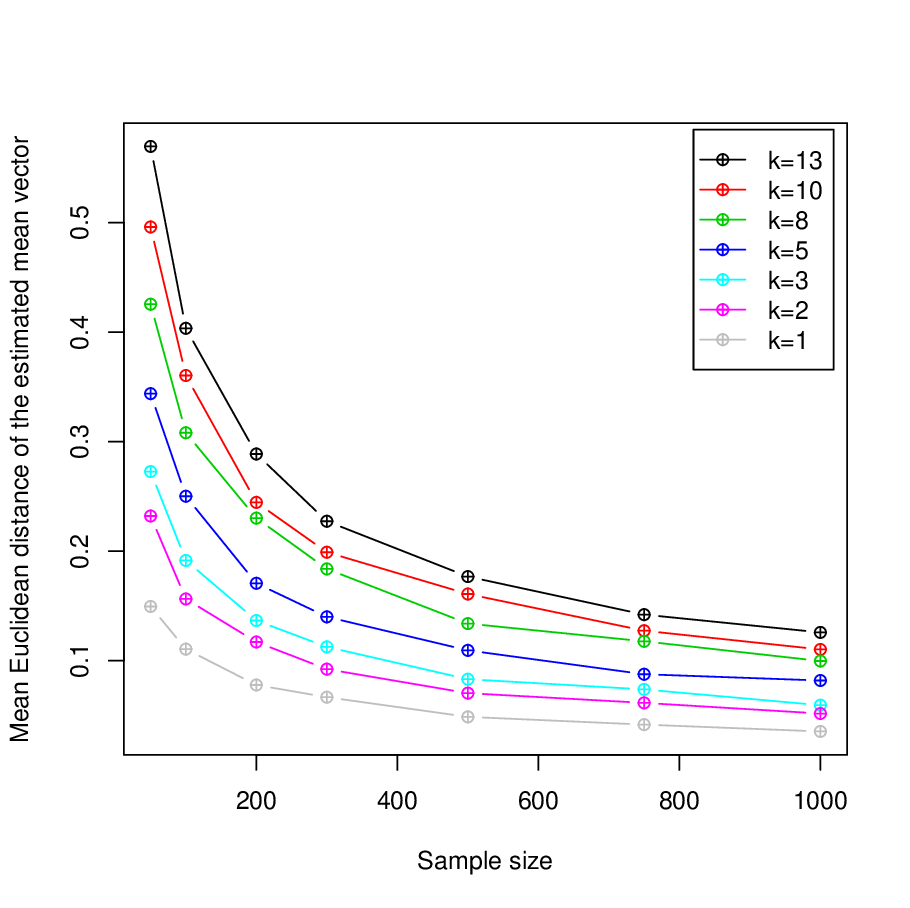} &
\includegraphics[scale=0.35,trim=50 0 0 30]{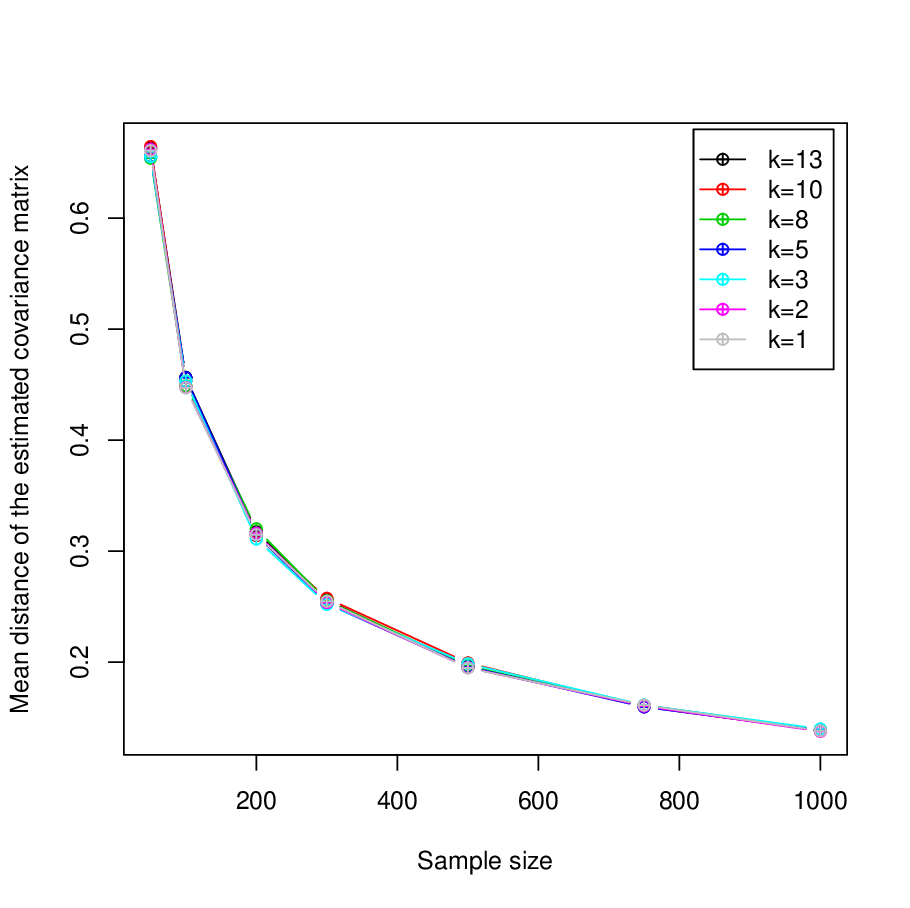} &
\includegraphics[scale=0.35,trim=50 0 0 30]{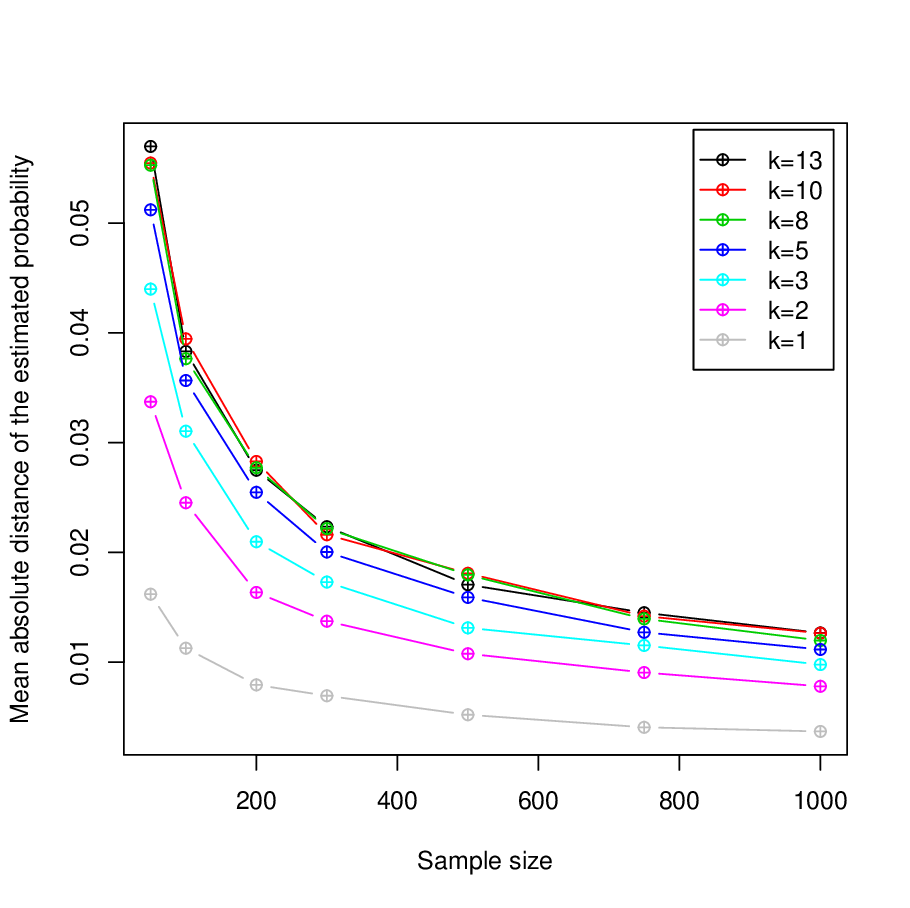} \\
\end{tabular}
\caption{All graphs contain the mean distance from each set of the parameters. The first column refers to the Euclidean distance of the estimated mean vector from the true mean vector. The second column refers to the mean distance between the estimated and the true covariance matrix. The third column refers to the mean absolute distance between the estimated probability and the true probability inside the simplex.}
\label{figure}
\end{figure}

Note that while we could have used the Kullback-Leibler divergence of the fitted multivariate normal from the true multivariate normal to evaluate the overall performance of our estimation method, we would then have had no individual information regarding the accuracy of our procedure in terms of estimating the probability left outside the simplex, the mean vector and the covariance matrix. The results of the simulation studies are presented in Table \ref{table1} and Figure \ref{figure}.

\begin{table}[h]
\begin{small}
\begin{center}
\begin{tabular}{l|c|c|c|c|c|c|c} 
\hline \hline
$\kappa$       & 0.5 & 1 & 2 & 3 & 5 & 7 & 10 \\ \hline \hline
$\alpha=-0.5$  & 0.0281 & 0.0925 & 0.1997 & 0.2800 & 0.3900 & 0.4630 & 0.5377 \\ \hline
$\alpha=0.5$   & 0.0223 & 0.1048 & 0.1849 & 0.3060 & 0.4217 & 0.4750 & 0.5356 \\ \hline \hline
\end{tabular}
\caption{Estimated probability left outside the simplex when $\alpha=-0.5$ and $\alpha=0.5$, calculated via Monte Carlo with 50,000,000 iterations. \label{table1} }
\end{center}
\end{small}
\end{table}

From Figure \ref{figure} we observe that when the probability left outside the simplex grows larger ($\kappa$ is larger), a larger sample size is required in order to get better estimates, for both the probability and the mean vector. The covariance matrix seems to be unaffected by the probability left outside the simplex. 

\subsection{Estimation of $\alpha$}
In the previous simulations, recall that the value of $\alpha$ was fixed. We now examine the performance of our estimation algorithm in relation to $\alpha$. \\

We focus on the large sample case in order to gain insight into the asymptotic behavior of $\hat{\alpha}$.  For this simulation study, we generated data as in Subsection \ref{sim1} with the mean vector set to
\begin{eqnarray*}
\pmb{\mu}=\left(1.715, 0.914, 0.115, 0.167\right)  
\end{eqnarray*} 
and the covariance matrices as in Equation (\ref{covk}).

For values of $\alpha$ ranging from $0$ up to $1$ with a step of $0.1$ we estimated these values for the different values of $\kappa$ using 4 sample sizes $(n = 1000, 5000, 10000, 20000)$. For each combination of $\alpha$, $\kappa$ and $n$ we used $1000$ repetitions. 

\begin{figure}[!ht]
\centering
\begin{tabular}{cc}
\includegraphics[scale=0.35,trim=0 20 0 20]{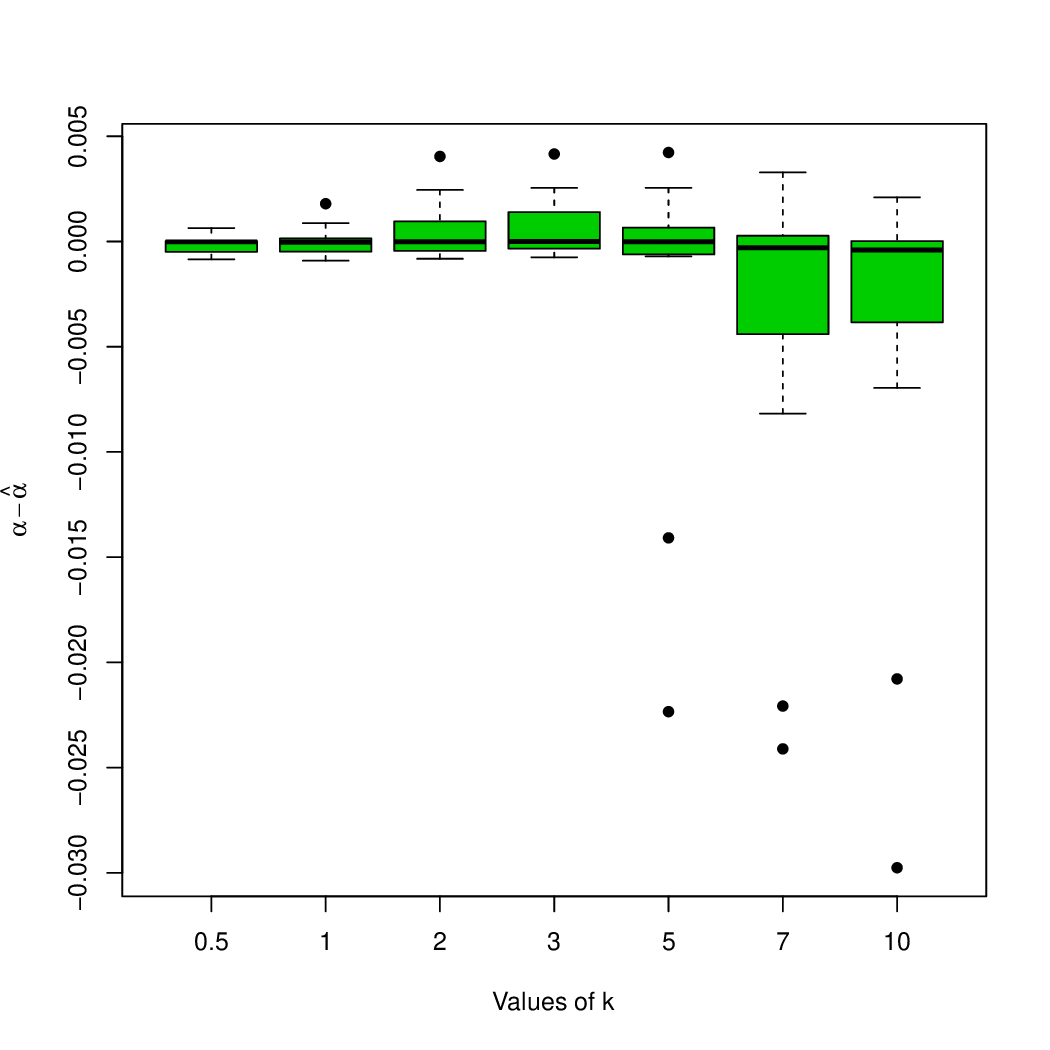} &
\includegraphics[scale=0.35,trim=0 20 0 20]{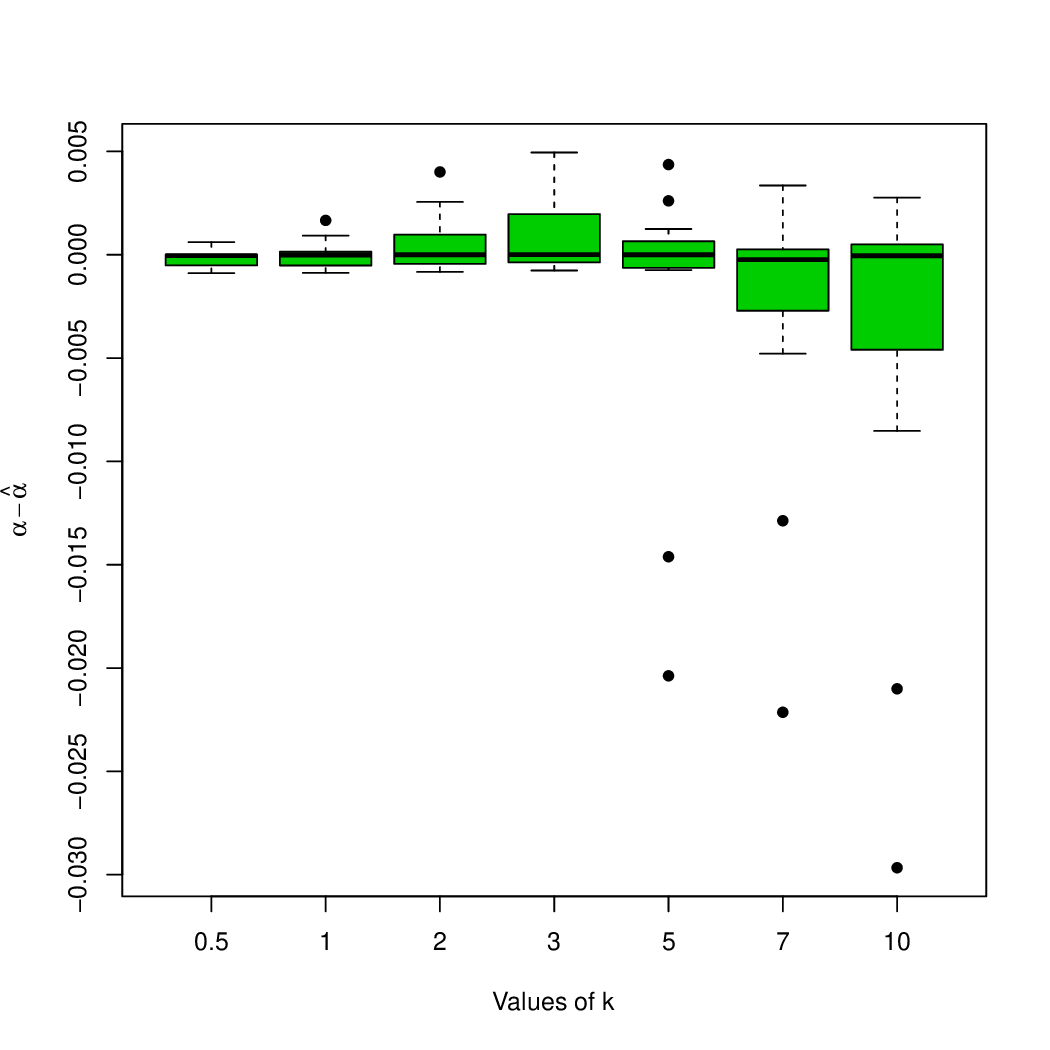}  \\
(a) n = 1000   &   (b)  n = 5000  \\
\includegraphics[scale=0.35,trim=0 20 0 20]{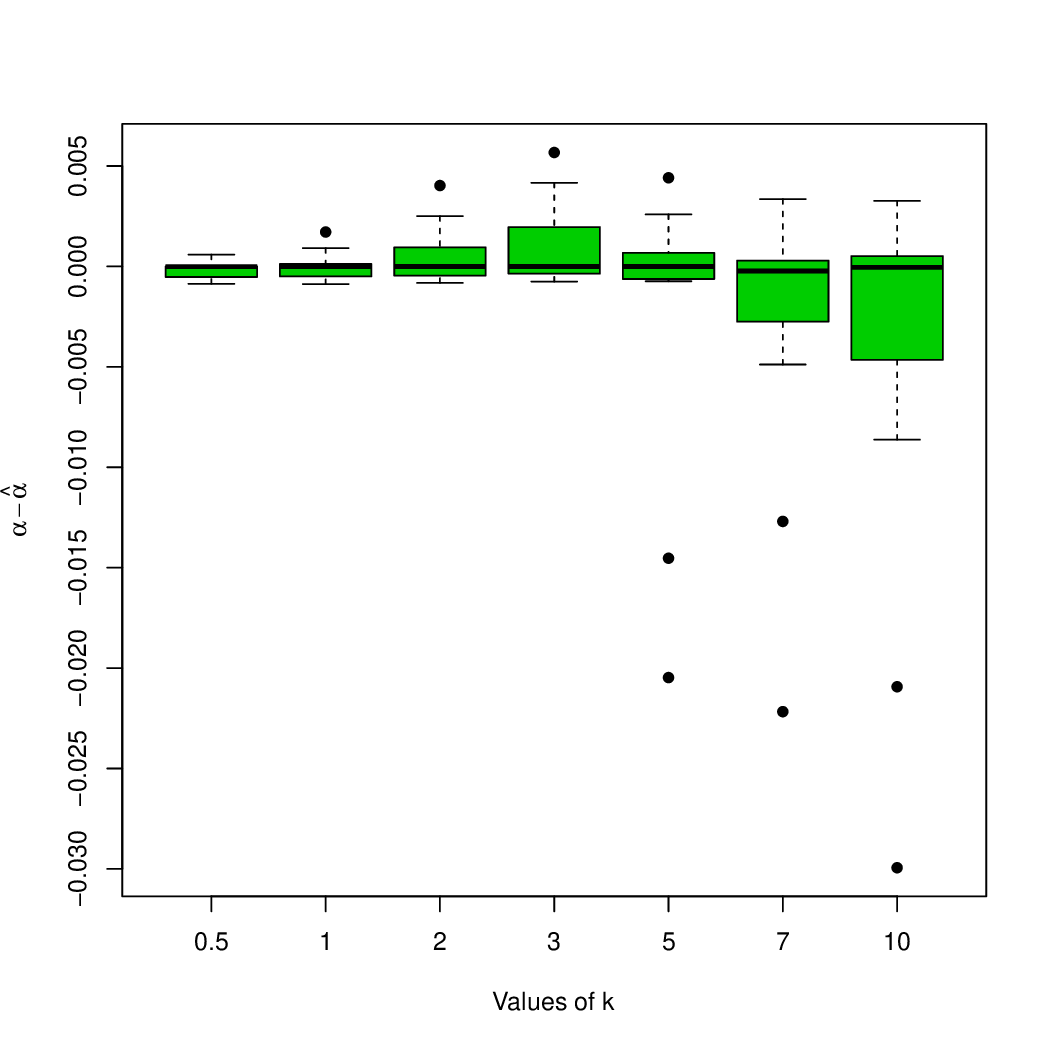} &
\includegraphics[scale=0.35,trim=0 20 0 20]{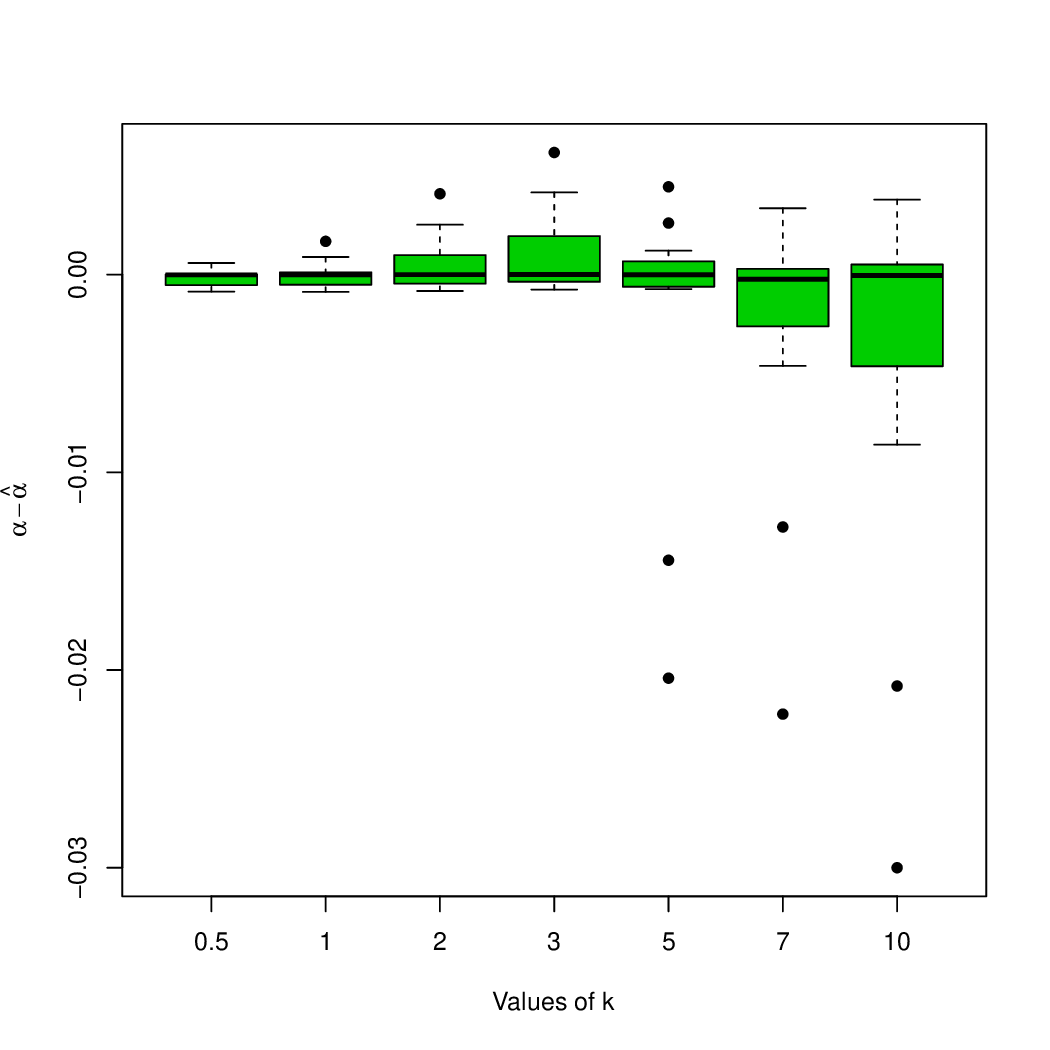}  \\
(c) n = 10000   &   (d)  n = 20000  
\end{tabular}
\caption{ Box plots of the range of $\alpha - \hat{\alpha}$ a a function of $\kappa$ for 4 different sample sizes. \label{figure2} }
\end{figure}

Figure \ref{figure2} shows the average bias of the $\alpha$ estimates in boxplots for each sample size. Each box corresponds to a value of $\kappa$ and is the average bias aggregated for all values of $\alpha$. For example, Figure \ref{figure2}(a) refers to a sample size equal to $1000$ and the first box contains information about the average biases of the $11$ values of $\alpha$. From the plots, as expected, the range in the variances increases with the value of $\kappa$, since higher values of $\kappa$ correspond to a higher probability of being left outside of the simplex. Table \ref{table3} presents $1-p$ for many combinations of values of $\alpha$ and $\kappa$ calculated using Monte Carlo simulation with $20,000,000$ repetitions. Clearly, $1-p$ increases as either $\kappa$ or $\alpha$ increases.

\begin{table}[h]
\begin{small}
\begin{center}
\begin{tabular}{c|c|c|c|c|c|c|c|c} 
\hline \hline 
               & \multicolumn{7}{c}{$\kappa$} \\ \hline \hline
$\alpha$    & 0.5 & 1 & 2 & 3 & 5 & 7 & 10 \\ \hline \hline
0.0  &  0.000  &  0.000  &  0.000  &  0.000  &  0.000  &  0.000  &  0.000  \\
0.1  &  0.000  &  0.000  &  0.000  &  0.000  &  0.000  &  0.000  &  0.002  \\
0.2  &  0.000  &  0.000  &  0.001  &  0.007  &  0.034  &  0.071  &  0.128  \\
0.3  &  0.000  &  0.004  &  0.040  &  0.091  &  0.187  &  0.316  &  0.348  \\
0.4  &  0.006  &  0.047  &  0.156  &  0.245  &  0.367  &  0.445  &  0.522  \\
0.5  &  0.043  &  0.149  &  0.306  &  0.402  &  0.516  &  0.583  &  0.648  \\
0.6  &  0.132  &  0.284  &  0.448  &  0.536  &  0.632  &  0.687  &  0.741  \\
0.7  &  0.258  &  0.423  &  0.571  &  0.644  &  0.722  &  0.768  &  0.812  \\
0.8  &  0.398  &  0.551  &  0.673  &  0.731  &  0.794  &  0.830  &  0.866  \\
0.9  &  0.535  &  0.661  &  0.757  &  0.802  &  0.851  &  0.880  &  0.907  \\
1.0  &  0.66   &  0.756  &  0.827  &  0.861  &  0.898  &  0.918  &  0.937  \\  \hline \hline
\end{tabular}
\caption{Probability left outside the simplex for many combinations of $\alpha$ and $\kappa$ values. \label{table3} }
\end{center}
\end{small}
\end{table}

\subsection{Estimation of the Computational Cost}
Using only $\alpha = 0.5$, we generated data as in Subsection \ref{sim1} with increasing sample sizes and, for each sample size, recorded the time (in seconds) required to estimate the true value of $\alpha$. The results are presented in Table \ref{table2}. As expected, the computational cost is mostly affected by $p$. For large sample sizes the computational burden is similar regardless of the probability of being outside of the simplex.

\begin{table}[h]
\begin{small}
\begin{center}
\begin{tabular}{c|c|c|c|c|c|c|c|c} 
\hline \hline
               & \multicolumn{7}{c}{$\kappa$} \\ \hline \hline
Sample size    & 0.5 & 1 & 2 & 3 & 5 & 7 & 10 \\ \hline \hline
50     &  0.052  &  0.058  &  0.065  &  0.075  &  0.095  &  0.124  &  0.148  \\
100    &  0.073  &  0.079  &  0.092  &  0.106  &  0.135  &  0.174  &  0.212  \\
200    &  0.090  &  0.089  &  0.096  &  0.105  &  0.130  &  0.162  &  0.195  \\
300    &  0.095  &  0.095  &  0.096  &  0.108  &  0.128  &  0.160  &  0.194  \\
500    &  0.084  &  0.083  &  0.091  &  0.100  &  0.123  &  0.156  &  0.188  \\
750    &  0.078  &  0.074  &  0.081  &  0.089  &  0.109  &  0.135  &  0.163  \\
1000   &  0.082  &  0.078  &  0.086  &  0.095  &  0.117  &  0.140  &  0.168  \\  
2000   &  0.267  &  0.390  &  0.357  &  0.352  &  0.339  &  0.276  &  0.302  \\
5000   &  0.638  &  0.933  &  0.838  &  0.841  &  0.816  &  0.675  &  0.722  \\ 
10000  &  1.435  &  2.113  &  1.905  &  1.915  &  1.858  &  1.544  &  1.723  \\ \hline \hline
\end{tabular}
\caption{Computational times (in seconds) required to estimate the value of $\alpha$, averaged over 1000 repetitions. \label{table2} }
\end{center}
\end{small}
\end{table}

\subsection{Estimation of Data Set Parameters}
in this simulation study we attempt to evaluate the performance of the EM algorithm for parameter values observed in practice by using the estimated parameters from the four data sets analyzed in Section \ref{data} as the true parameters,  To accomplish, we generated compositional data using the proposed folded multivariate normal distribution (with the same number of dimensions and sample sizes as the real data) with estimated parameters $\hat{\alpha}$,  $\hat{p}$, $\pmb{\hat{\mu}}_{\alpha}$  and $\hat{\pmb{\Sigma}}_{\alpha}$ .  For each data set, $1000$ samples were simulated and we calculated the average distances of the estimated parameters from the true parameters, as described in Subsection \ref{sim1}. Table \ref{datasim} shows the results of this Monte Carlo study. The large bias observed for $\alpha$ with the Sharp II data set is perhaps to be expected given the poor fit that was observed in Subsection \ref{SharpII}. For Example 4 (\textit{Labor Force}), the bias tends to be larger for $\hat{p}$, $\pmb{\hat{\mu}}_{\alpha}$  and $\hat{\pmb{\Sigma}}_{\alpha}$ compared to the other data sets.

\begin{table}[h]
\begin{small}
\begin{center}
\begin{tabular}{l|c|c|c|c|c|c} 
\hline \hline
          & \multicolumn{6}{c}{Estimated bias}  \\ \hline
data set   & $n$ & $D$ & $\alpha$  &  $p$  & $\pmb{\mu}_{\alpha}$  & $\pmb{\Sigma}_{\alpha}$ \\ \hline \hline
Sharp I  & $25$ & $3$ & 0.054 & 0.042 &  0.248 &  0.633 \\  \hline
Sharp II & $25$ & $3$ &  0.219  & 0.013  & 0.145 & 0.594 \\ \hline
Coffee & $30$ & $6$ & 0.154 & 0.036 & 0.160  &  1.191 \\ \hline
Labor Force  & $124$ & $6$ & 0.049 & 0.510 & 1.257  &  1.628 \\ \hline \hline
\end{tabular}
\caption{Estimated bias of the parameters in the real data sets using Monte Carlo.} \label{datasim} 
\end{center}
\end{small}
\end{table}

\subsection{Comparison of the folded and simple $\alpha$-transformations}
We will now illustrate the effect of the folding transformation on the 4 real data sets. For each data set we generated data from the $\alpha$-folded multivariate normal distribution using the estimated parameters as the true parameters, but with some modifications to induce various values of $p$. Specifically for the two Sharp's artificial data sets, we generated data from the estimated parameters but multiplied all the elements of the covariance matrix by $\kappa$, for various values of $\kappa$. 
Similarly, for the data sets \textit{Coffee Aroma} and \textit{Labor Force} we first multiplied all the elements of the mean vector by different values of $\lambda$ and then generated data. For both data sets, the values of $\kappa$ and $\lambda$ affect the probability left outside the simplex. For every generated data set we estimated the parameters of the $\alpha$-folded multivariate normal distribution and of the $\alpha$-normal (Tsagris et al, 2011). In both cases we generated $1000$ compositional vectors.  

\begin{figure}[!ht]
\centering
\begin{tabular}{cc}
\includegraphics[scale=0.45]{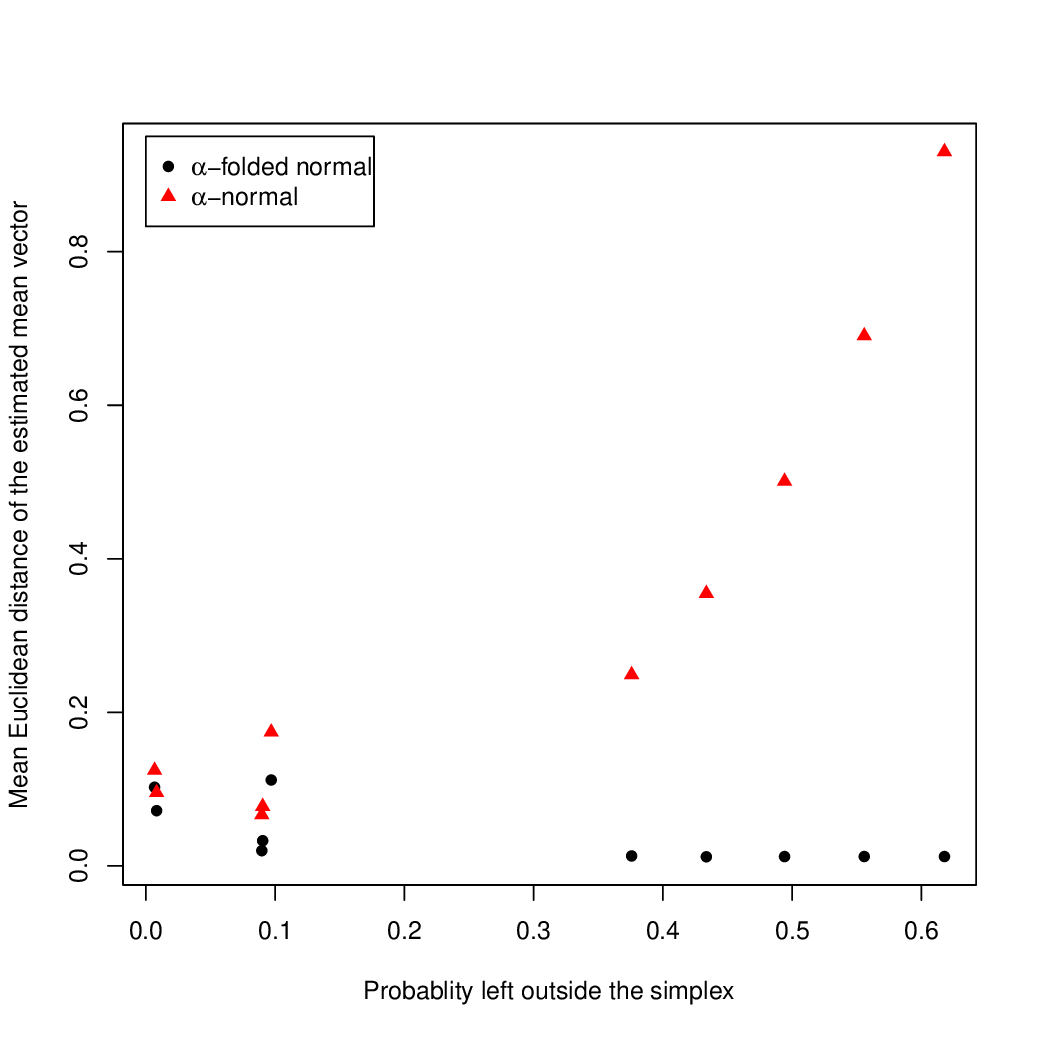} &
\includegraphics[scale=0.45]{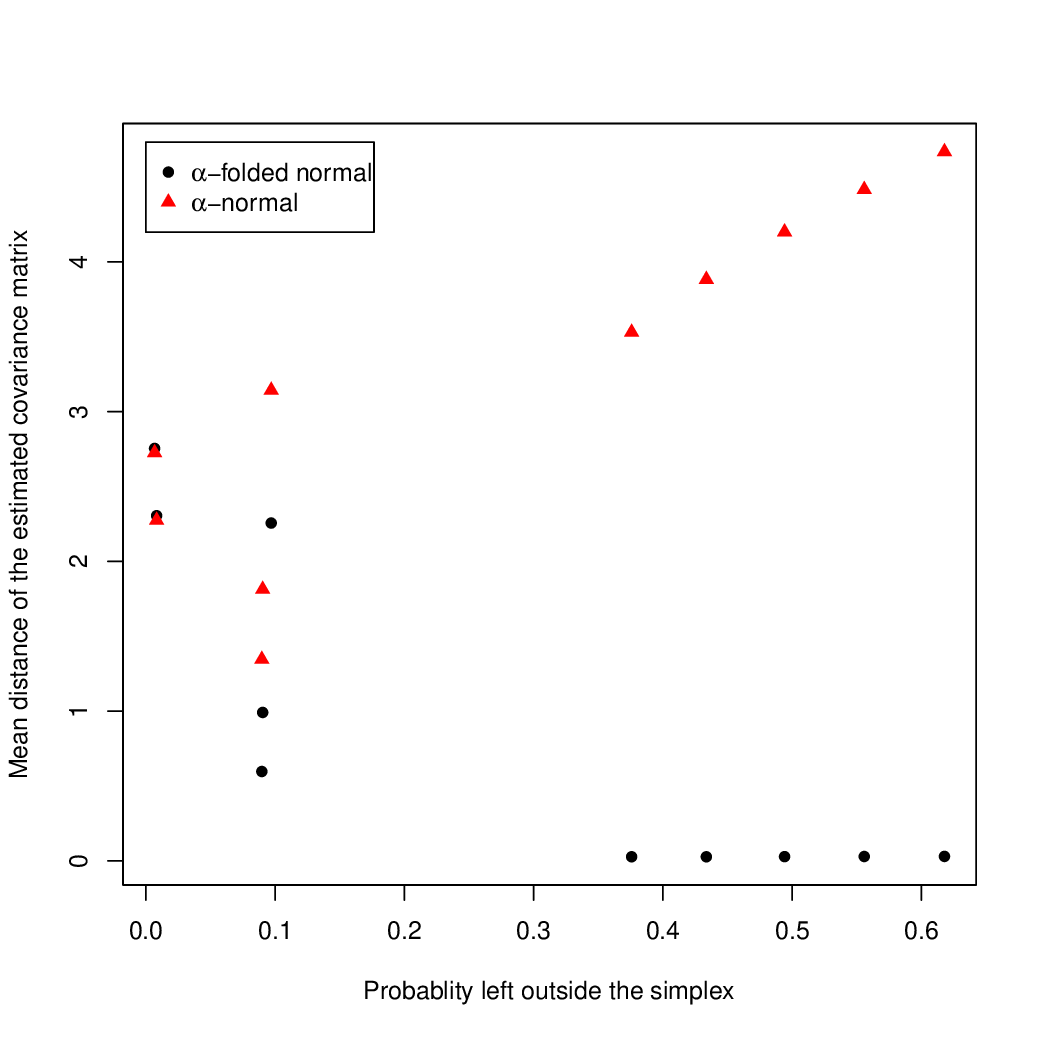}  \\
\multicolumn{2}{c}{ \includegraphics[scale=0.45]{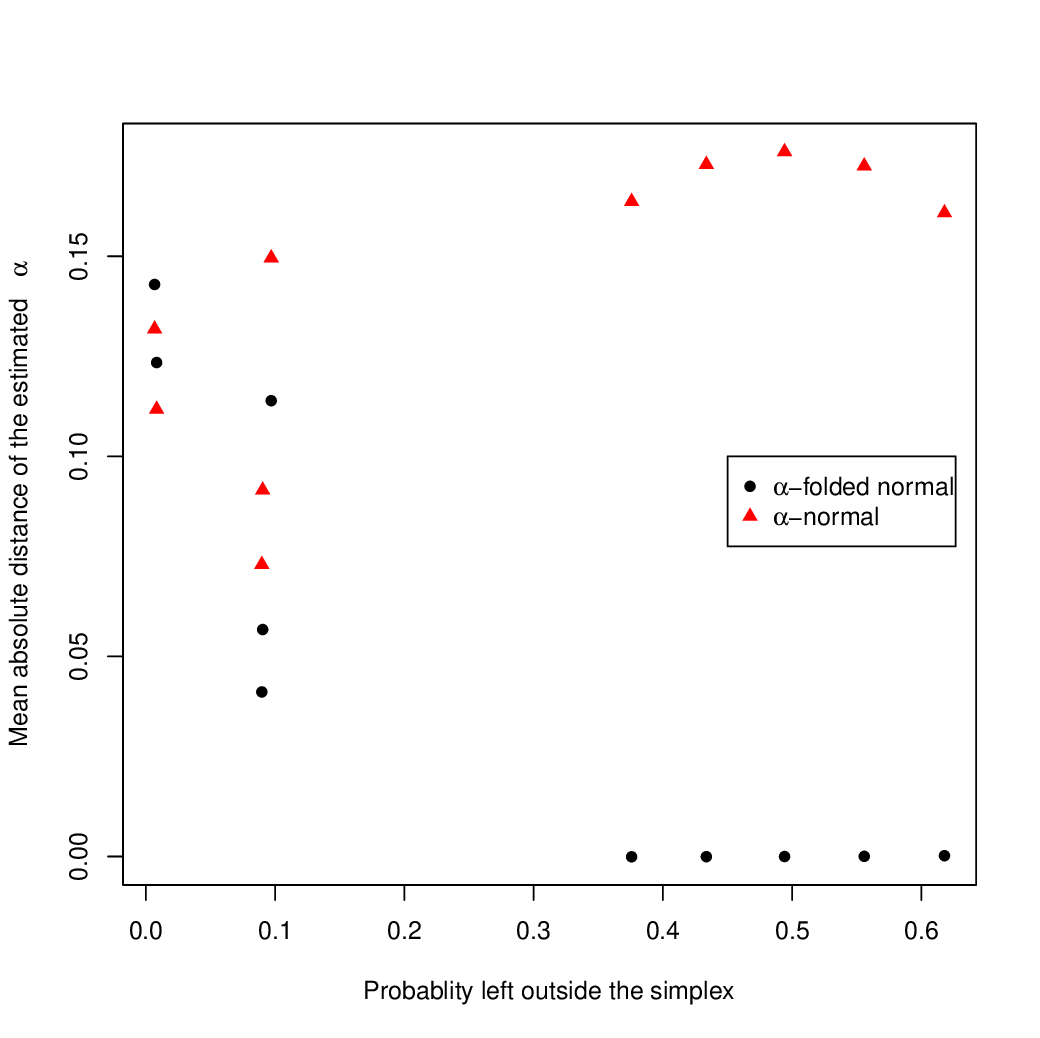} }
\end{tabular}
\caption{Mean estimated distances (for the mean vector, covariance matrix and values of $\alpha$) between the true and estimated parameters of the $\alpha$-folded multivariate normal distribution and of the multivariate normal distribution after applying the $\alpha$-transformation for a range of probabilities left outside the simplex. \label{ha} }
\end{figure}

The accuracy of the mean vector was evaluated using the Euclidean distance and for the estimation of the covariance matrix the metric defined in (\ref{covdist}) was used. The absolute difference between the true and the estimated value of $\alpha$ was also calculated. The results for the two Sharp's artificial data sets were similar and hence we only present the results for the first data set in Figure \ref{ha}. Table \ref{table_coffee} provides the results for the two real data sets, namely \textit{Coffee Aroma} and \textit{Labor Force}.

Figure \ref{ha} depicts the effect of the probability left outside the simplex. When this probability is considerably high, the $\alpha$-normal fails to fit the data adequately and the estimates are highly biased.  The results in Table \ref{table_coffee} also suggest that the bias in the estimates decreases often substantially for the $\alpha$-folded normal model as the probability left outside the simplex increases, but increases (though sometimes only slightly) for the $\alpha$-model.

\begin{table}[h]
\begin{small}
\begin{center}
\begin{tabular}{c|cc|cc|cc} 
\hline \hline
Probability    &  \multicolumn{2}{c}{Euclidean distance}  &  \multicolumn{2}{c}{Covariance}  &  \multicolumn{2}{c}{Absolute difference} \\ 
left outside &  \multicolumn{2}{c}{of the mean vector}  &  \multicolumn{2}{c}{matrix metric}  & \multicolumn{2}{c}{for $\alpha$}   \\
the simplex        &  \multicolumn{6}{c}{}   \\ \hline \hline
       &   \multicolumn{6}{c}{Coffe Aroma Data}   \\ \hline   
       &  $\alpha$-folded normal & $\alpha$-normal &  $\alpha$-folded normal & $\alpha$-normal &  $\alpha$-folded normal & $\alpha$-normal \\
              \hline
0.018  &  0.411  &  0.438  &  1.327  &  1.319  &  0.130  &  0.102  \\
0.162  &  0.562  &  0.746  &  1.619  &  2.202  &  0.101  &  0.096  \\
0.538  &  0.168  &  0.763  &  0.395  &  2.275  &  0.026  &  0.094  \\
0.642  &  0.072  &  0.778  &  0.087  &  2.345  &  0.008  &  0.092   \\ \hline \hline
       &   \multicolumn{6}{c}{Labor Force Data}   \\ \hline   
       &  $\alpha$-folded normal & $\alpha$-normal &  $\alpha$-folded normal & $\alpha$-normal &  $\alpha$-folded normal & $\alpha$-normal \\
0.246  &  1.277  &  1.619  &  1.515  &  2.016  &  0.017   &  0.024 \\
0.586  &  0.729  &  1.668  &  0.939  &  2.014  &  -0.031  &  0.026 \\
0.701  &  0.422  &  1.716  &  0.564  &  2.011  &  -0.035  &  0.029  \\ \hline \hline
\end{tabular}
\caption{Estimated differences between the estimated and the true parameters. The $\alpha$-folded normal is the model proposed in this paper, whereas the $\alpha$-normal refers to the model proposed by Tsagris et al. (2011). \label{table_coffee} }
\end{center}
\end{small}
\end{table}

\subsection{Estimation in Higher Dimensions}
\begin{figure}[!ht]
\centering
\begin{tabular}{cc}
\includegraphics[scale=0.42,trim=0 20 0 20]{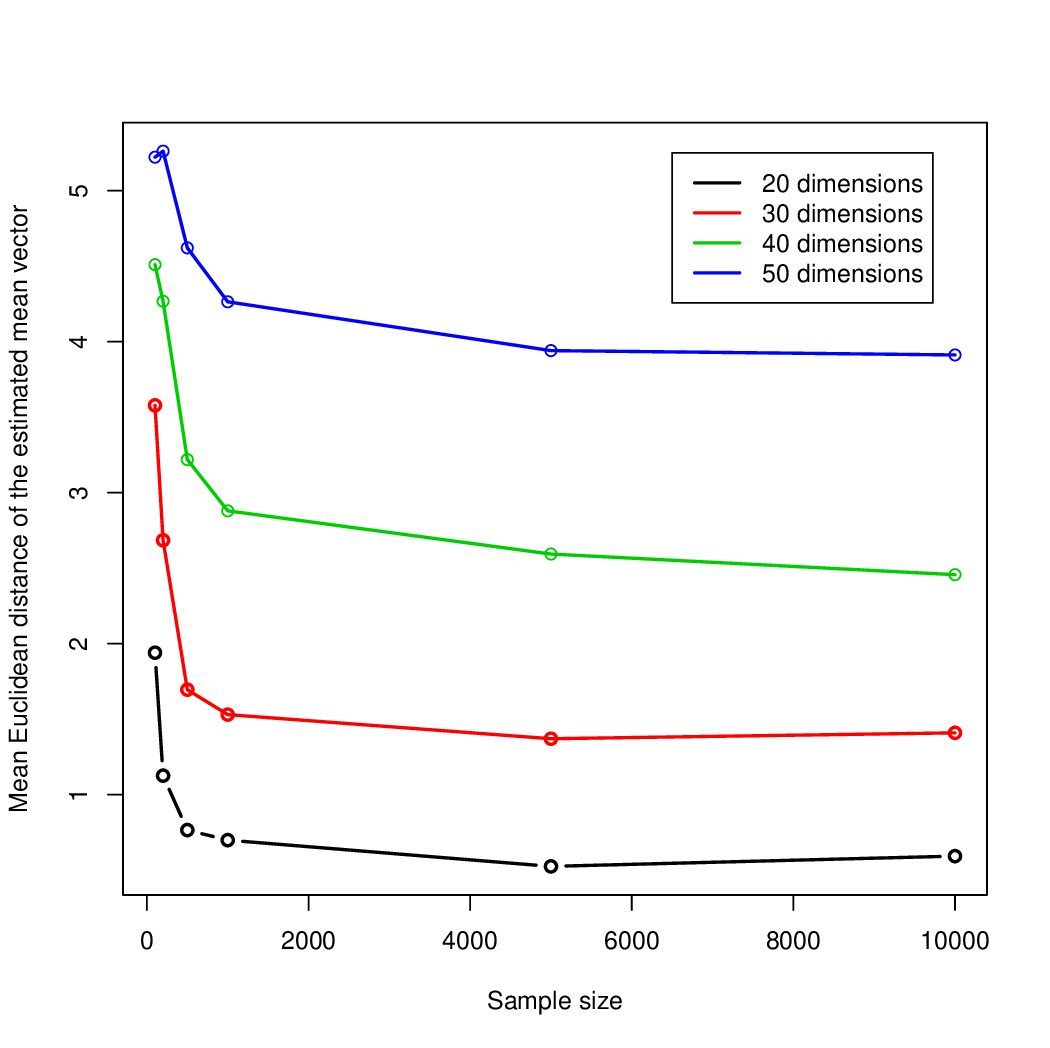} &
\includegraphics[scale=0.42,trim=0 20 0 20]{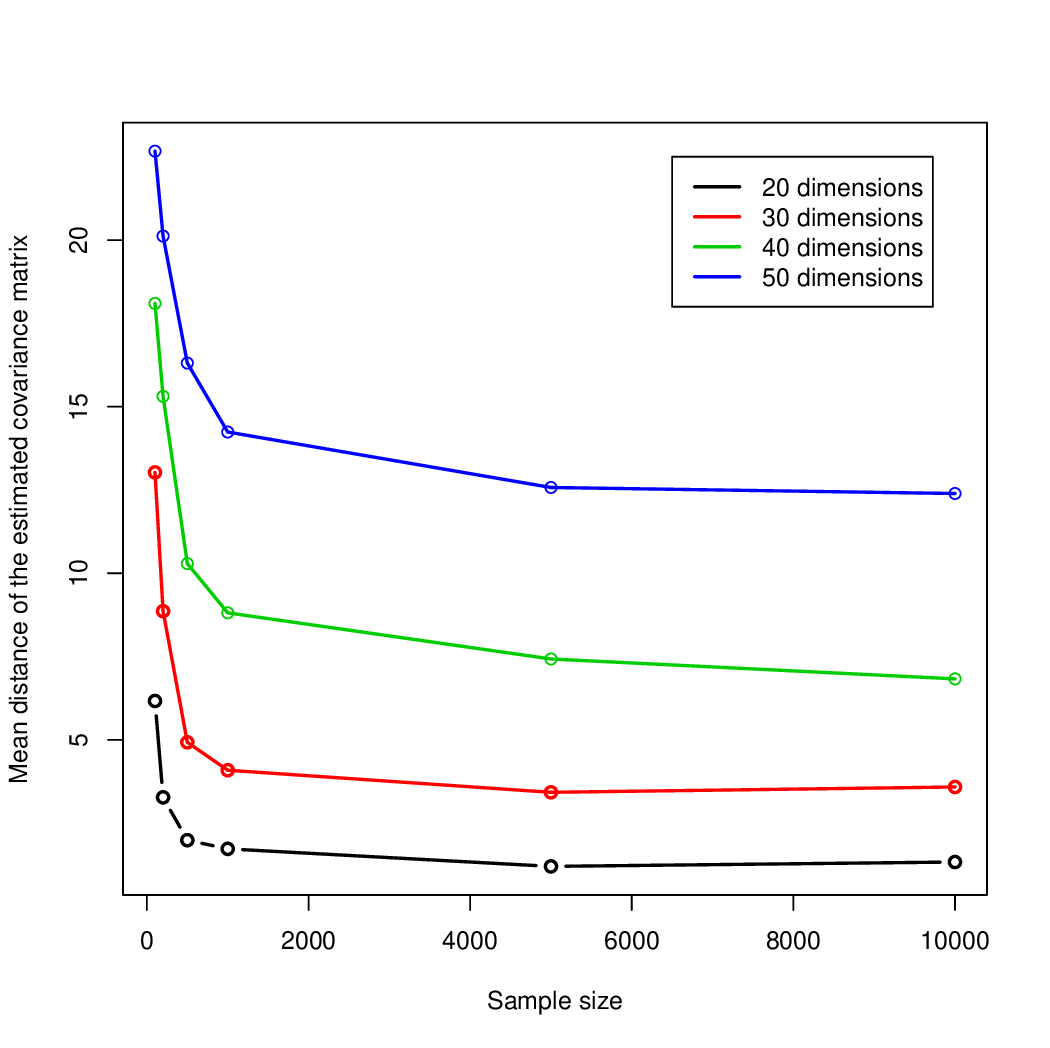} \\
\includegraphics[scale=0.42,trim=0 20 0 20]{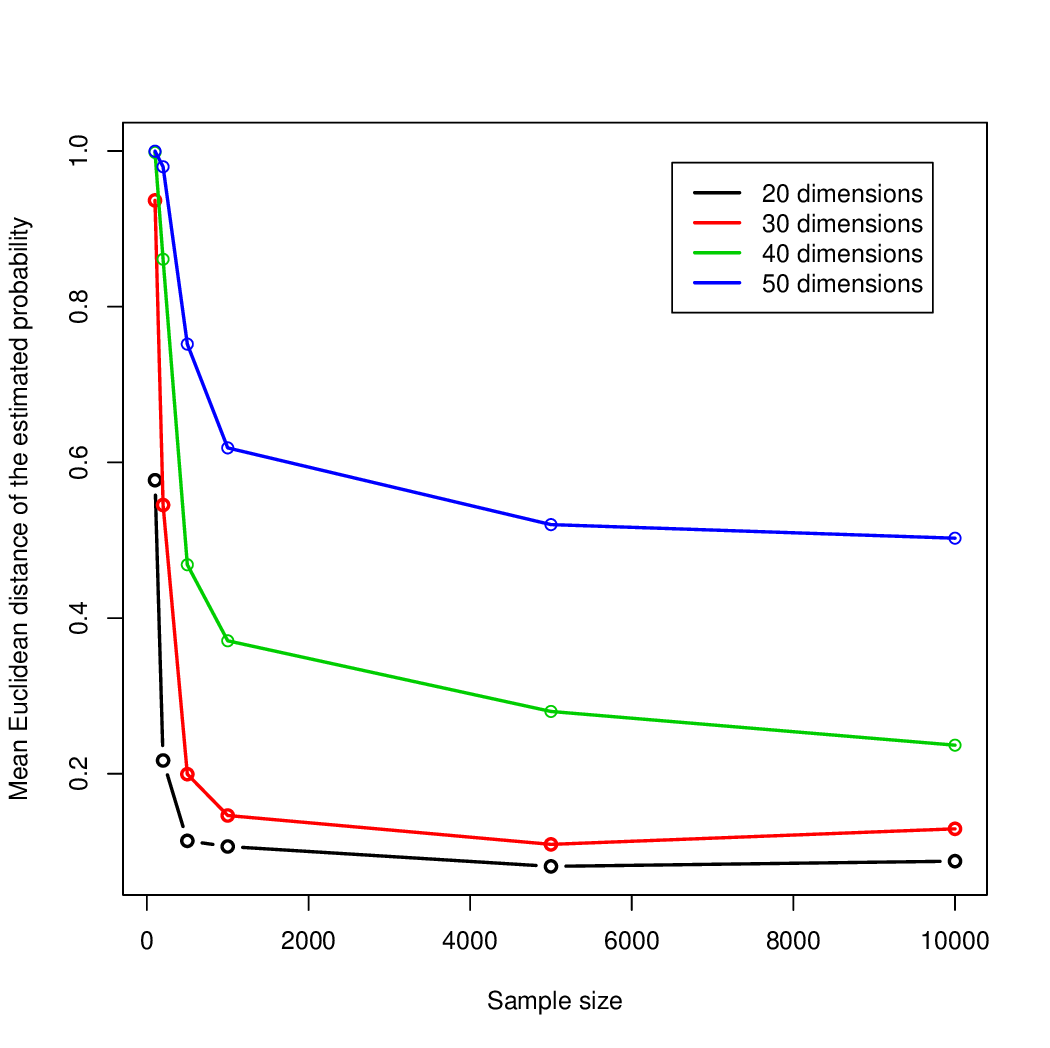}  &
\includegraphics[scale=0.42,trim=0 20 0 20]{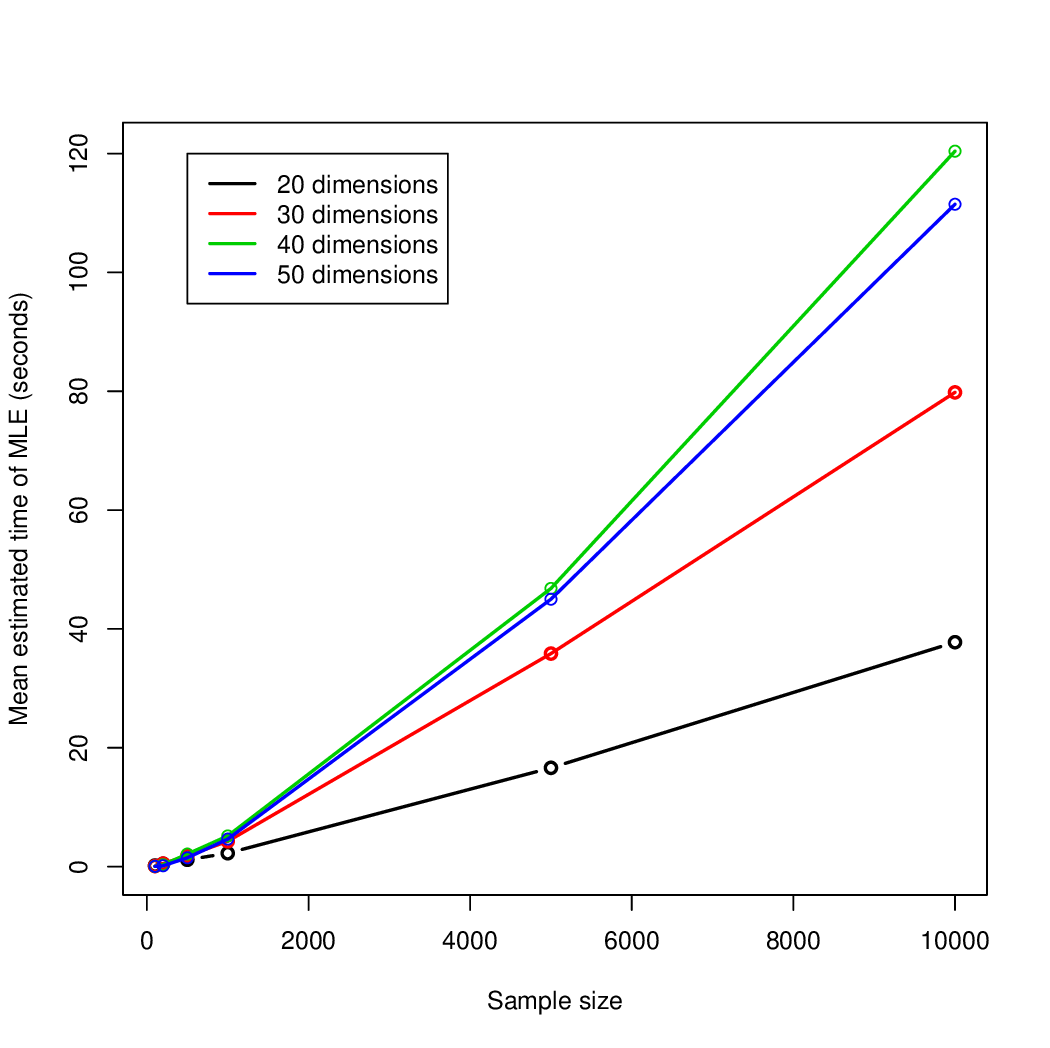}  
 \end{tabular}
\caption{Mean estimated distances (for the mean vector, covariance matrix and probability left outside the simplex) between the true and estimated parameters of the $\alpha$-folded multivariate normal distribution when $\alpha=0.5$ for a range of sample sizes. The plot at the bottom right refers to the estimated computational time (in seconds) required by the EM algorithm.\label{dims} }
\end{figure}

\begin{figure}[!ht]
\centering
\includegraphics[scale=0.45,trim=0 20 0 20]{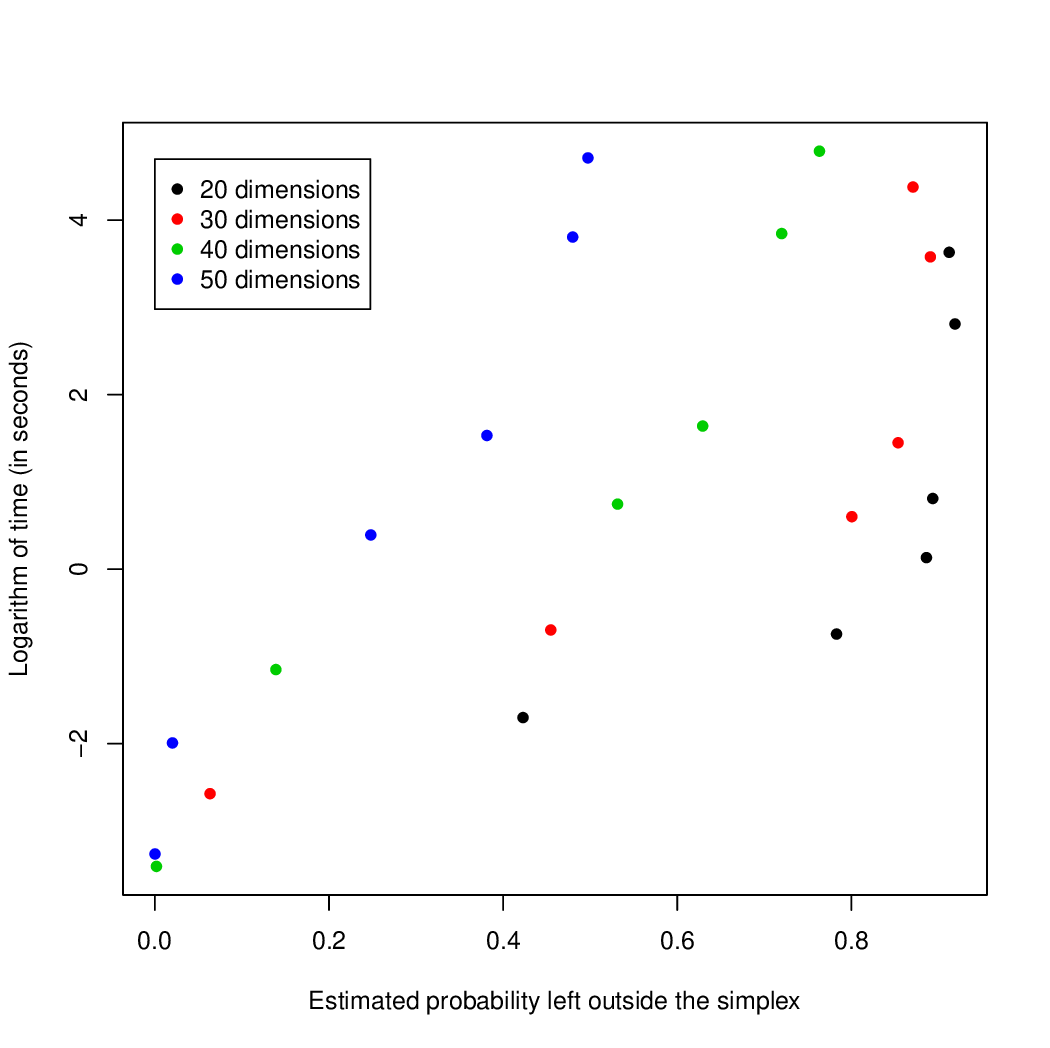}  
\caption{Computational time (logarithm of the seconds) required by the EM algorithm versus the estimated probability left outside the simplex, for different number of components and a range of sample sizes. \label{pd_ti} }
\end{figure}

The case of high dimensional compositional data was only recently examined (Lin et al., 2014; Fang et al., 2015; Shi et al., 2016; Cao, Lin \& Li, 2018a, 2018b). In line with the direction of these papers, we also examined the performance of the $\alpha$-folded model with higher dimensions for a fixed value of $\alpha$. We simulated data for various sample sizes $n = (100, 200, 500, 1000, 5000, 10000)$ and a varying number of components $D = (10, 20, 30, 40, 50)$. We set the value of $\alpha$ equal to $0.5$, and generated data from a multivariate normal distribution with a mean vector generated from a standard normal and a diagonal covariance matrix with the variances generated from an exponential with mean 2. Like in all previous cases, the generated vectors are mapped into the simplex using Equation (\ref{trans}). We calculated the Euclidean distance between the true and the estimated mean vectors, the discrepancy between the true and the estimated covariance matrix using Equation (\ref{covdist}), the absolute differences between the true and the estimated values of $p$ and, finally, the computational time required by the EM algorithm.  Results are shown in Figures \ref{dims} and \ref{pd_ti}.

In terms of how accurately the parameters can be estimated for higher dimensional data, not surprisingly, the bias tends to be larger as the dimension increases but little improvement is observed for samples sizes larger than $1000$.  The results suggest that our estimation algorithm may not be reliable for dimensions larger than 30 and that for dimensions between 20 and 30, a large sample size is required.  From Figure \ref{pd_ti}, the estimated probability left outside the simplex affects the computational time but the dimension size appears to be less important.

\section{Conclusions}  \label{conclusion}
In this paper we developed a novel parametric model, with nice properties, for compositional data analysis. The model is an extension of the model proposed by Tsagris, Preston \& Wood (2011) in which the $\alpha$-transformation and corresponding multivariate normal distribution was introduced. A drawback of their model is that it does not take into account the probability left outside the simplex, and this deficiency motivated the development of our proposed folded model.  Simulation study results suggest that if the probability left outside the simplex is large (as it was for one of the real-life examples), parameter estimates using the model in Tsagris, Preston \& Wood (2011) tend to be quite biased.  While we dealt with the probability left outside the simplex through folding, another possible solution would be to use truncation \citep{dobigeon2007}.

The proposed model is also an extension of the popular logistic normal distribution which corresponds to $\alpha = 0$.  In the results that were presented, the $\alpha$-folded model appeared to fit the data adequately when the logistic normal distribution did not. This is inline with other work in which the log-ratio transformation failed to capture the variability of the data. See, for example, Tsagris, Preston \& Wood (2011), Baxter (2006) and Sharp (2006).

The use of a multivariate model other than the multivariate normal distribution, such as the multivariate skew normal distribution (Azzalini \& valle, 1996) has also been suggested. The challenge, however, with this distribution is that more parameters need to be estimated, thus making the estimation procedure more difficult because the log-likelihood has many local maxima. Another, perhaps simpler, alternative model is the multivariate $t$ distribution. Bayesian analysis and regression modeling are two suggested research directions. 

As previously mentioned and similar to the Box-Cox transformation, zero values are not compatible with our proposed model. Note that the zero issue also arises with the logistic normal distribution. However, it is possible to generalize most of the analyzes suggested for the logistic normal distribution using our proposed folded model, including extensions that allow zeros.

As with the ilr transformation, the $\alpha$-transformed data have no clear interpretation. The same issue occurs with the additive log-ratio transformation $\left(\left\lbrace\log{\left(x_i/x_D \right)} \right\rbrace_{i=1,\ldots,D-1} \right)$ (Aitchison, 2003) and while the mean vector and covariance matrix are interpretable, they depend on the denominator component. Hence among models for compositional data, there is a trade-off between interpretability and better fit. Presumably the application would dictate which is of greater importance.

As is standard practice in log-ratio transformation analysis, if one is willing to exclude from the sample space the boundary of the simplex, which includes observations that have one or more components equal to zero, then the folded $\alpha$-transformation (\ref{alpha}) and its inverse are well defined for all $\alpha \in \mathbb{R}$, and the corresponding $\alpha$-folded model provides a new approach for the analysis of compositional data with the potential to provide an improved fit over traditional models.

\section*{Appendix}
\begin{appendix}
\renewcommand{\theequation}{A.\arabic{equation}}
\renewcommand\thefigure{A.\arabic{figure}}    
\setcounter{equation}{0}  
\setcounter{figure}{0}  

\subsection*{The Helmert sub-matrix}
The Helmert matrix is a $D \times D$ orthogonal matrix. The Helmert sub-matrix has the first row omitted, hence is a $D- 1 \times D$ matrix, the structure of which is presented below. 

\begin{eqnarray}
{\bf H}_{d, d + 1}= 
\left( \begin{array}{cccccc}
\frac{1}{\sqrt{2}}                 & -\frac{1}{\sqrt{2}} &  0                  & \ldots & \ldots      & 0       \\
\frac{1}{\sqrt{2}}                 & \frac{1}{\sqrt{6}}  & -\frac{2}{\sqrt{6}} & 0      & \ldots      & \vdots  \\
\vdots                             & \vdots              &  \vdots             & \ddots & \vdots      & \vdots  \\
\frac{1}{\sqrt{i\left(i+1\right)}} & \ldots              & \ldots              & \frac{1}{\sqrt{i\left(i+1\right)}} & 
-\frac{i}{\sqrt{i\left(i+1\right)}} & 0 \\  
\vdots                             & \vdots                             & \vdots              & \vdots              & \vdots & 0       \\
\frac{1}{\sqrt{\left(D-1\right)D}} & \ldots                             & \ldots              & \ldots  & \frac{1}{\sqrt{\left(D-1\right)D}} &         
-\frac{D-1}{\sqrt{\left(D-1\right)D}}  \\   
\end{array} \right) 
\end{eqnarray}

\subsubsection*{Explanation of the folding transformation}

We will show that if $\mathbf{y}\in \mathbb{R}^{D-1}\setminus\mathbb{A}^{D-1}$ then Equation (\ref{foldin}) transforms $\mathbf{y}$ from $\mathbb{R}^{D-1}\setminus\mathbb{A}^{D-1}$ to $\mathbb{S}^{D-1}$.  We will consider only the case that $\alpha >0$ as the case for $\alpha <0$ is similar.

Suppose $\mathbf{y}\in\mathbb{R}^{D-1}\setminus\mathbb{A}^{D-1}$ and let $\mathbf{w}= \mathbf{H}^T\mathbf{y}$.  Then $\mathbf{w}\in \mathbb{Q}_0$ from Equation (\ref{ilr}) and properties of $\mathbf{H}$.  Also, for $\alpha \neq 0$, $\mathbf{w} \notin \mathbb{Q}_\alpha^{D-1}$ and this implies that there exists a $w_i < -1/\alpha$ (in which case, $\min(\mathbf{w}) <-1/\alpha$) or a $w_i>(D-1)/\alpha$ (in which case, $\max(\mathbf{w}) >(D-1)/\alpha$).  Note, however, that we need only to consider the case in which $\min(\mathbf{w}) <-1/\alpha$ because if $\min(\mathbf{w})\nless-1/\alpha$, we can show that $\max(\mathbf{w}) \leq (D-1)/\alpha$.  Specifically,
assume that $\min(\mathbf{w}) \geq -1/\alpha \Rightarrow w_i\geq -1/\alpha\  \forall i = 1,\ldots D$. If we let $w_{D} = \max(\mathbf{w})$, we have that 
\begin{eqnarray}
\label{proofmax}
-\sum_{i=1}^{D-1}\frac{1}{\alpha} \leq \sum_{i=1}^{D-1}w_{i} \Rightarrow 
-\frac{D-1}{\alpha} \leq \sum_{i=1}^{D-1}w_{i} = -w_{D} \Rightarrow w_{D} \leq \frac{D-1}{\alpha},
\end{eqnarray}
since $\mathbf{w} \in \mathbb{Q}_0$ and $\sum_i^{D}w_i=0$.

Assume $\min(\mathbf{w}) <-1/\alpha$ and define 
\begin{equation}
\label{wprimealpha}
\mathbf{w}^{\prime\alpha} = \frac{\mathbf{w}}{\alpha|\min(\mathbf{w})|.} 
\end{equation}
Since $\sum_i^{D}w_i=0$, at least one component must be negative (if not, all components are zero) and if we divide by the absolute value of the smallest component, it is straightforward to show that $\min(\mathbf{w}^{\prime\alpha}) = -1/\alpha$. Note that if Equation (\ref{winv}) is applied to $\mathbf{w}^{\prime\alpha}$, $\mathbf{y}$ is transformed to the boundary of the simplex (that is, one component of the resulting vector will be zero). 

Let $q^* = \alpha \min(\mathbf{w})$.  To transform $\mathbf{w}$ inside $\mathbb{Q}_\alpha^{D-1}$ (rather than on the boundary), consider 
\[\mathbf{w}^\alpha = \frac{\mathbf{w}}{q^{*2}}.\] 
We need to show that $\mathbf{w}^\alpha \in \mathbb{Q_\alpha}^{D-1}$ since in this case, we can apply Equation (\ref{winv}) to transform $\mathbf{w^\alpha}$ to the simplex.  Clearly $\sum_{i=1}^D \mathbf{w}^\alpha_i=0$ so we need to show that 
$-1/\alpha < \mathbf{w}_i^\alpha<(D-1)/\alpha$.  

Since $\min{(\bf w)}< -\frac{1}{\alpha} 
\Rightarrow q^* < -1 \Rightarrow q^*2 >1$.  Therefore $\min \left ( \frac{\mathbf{w}}{q^{*2}} \right ) > -1/\alpha$. To show that $\max(\mathbf{w}) <(D-1)/\alpha$, we can simply argue as we did in Equation (\ref{proofmax}) that $\min \left ( \frac{\mathbf{w}}{q^{*2}} \right ) > -1/\alpha \Rightarrow \max \left (\frac{\mathbf{w}}{w^{*2}} \right ) < (D-1)/\alpha$. Therefore, $\mathbf{w}^\alpha \in \mathbb{Q_\alpha}^{D-1}$ and can be transformed inside the simplex via Equation (\ref{winv}). 



\subsubsection*{Proof of Lemma 3.1}
Let us begin by deriving the Jacobian determinant of (\ref{stayalpha}) at first. The map (\ref{stayalpha}) is degenerate due to the constraints $\sum_{i=1}^Dx_i=1$ and $\sum_{i=1}^Du_i=1$. In order to make (\ref{alpha}) non-degenerate we consider the version of (\ref{stayalpha}) as follows
\begin{eqnarray} \label{alpha2}
\mathbf{u}_a\left\lbrace\left(x_i\right) \right\rbrace=\frac{x_i^{\alpha}}{\sum_{j=1}^{D-1}x_j^{\alpha}+\left(1-\sum_{j=1}^{D-1}x_j \right)^{\alpha}} \ \ \ i=1,\ldots,d.
\end{eqnarray}

The (\ref{alpha2}) is presented to highlight that in fact we have $d=D-1$ and not $D$ variables. 

Let us start by proving the Jacobian of (\ref{stayalpha}) or (\ref{alpha2}). We denote $S(\alpha)=\sum_{j=1}^Dx_j^{\alpha}$, where $x_D=1-\sum_{j=1}^{D-1}x_j$. The diagonal and the non-diagonal elements of the Jacobian matrix are as follows. 
 
\begin{eqnarray*}
\frac{du_i}{dx_j}=
\left\lbrace
\begin{array}{cc}
\frac{\alpha x_i^{\alpha-1}S(\alpha)-x_i^{\alpha}(\alpha x_i^{\alpha-1}-\alpha x_D^{\alpha-1})}{S^2(\alpha)} & i = j \\ 
\-\frac{x_i^{\alpha}\left(\alpha x_j^{\alpha-1}-\alpha x_D^{\alpha-1}\right)}{S^2(\alpha)} \ \ \left(i \neq j \right) & i \neq j
\end{array}
\right\rbrace
\end{eqnarray*}
The Jacobian takes the following form (Mardia, Kent \& Bibby, 1979): 
\begin{eqnarray*}
\left|{\bf J}\right|=\left|{\bf A}-{\bf BC}^T\right|S^{-2(D-1)}(\alpha)=\left|{\bf A}\right|(1-{\bf C}^T{\bf A}^{-1}{\bf B})S^{-2(D-1)}(\alpha),
\end{eqnarray*}
where ${\bf A}$ is a diagonal $(D-1)\times (D-1)$ matrix with elements $\alpha x_i^{\alpha-1}S(\alpha)$ and ${\bf B}$ and ${\bf C}$ are defined as 
\begin{eqnarray*}
{\bf B}=\left(x_1^{\alpha}, \ldots, x_{D-1}^{\alpha}\right)^T \ \ \text{and} \ \ 
{\bf C}=\alpha\left(x_1^{\alpha-1}-x_D^{\alpha-1}, \ldots, x_{D-1}^{\alpha-1}-x_D^{\alpha-1}\right)^T.
\end{eqnarray*}
Then
\begin{eqnarray*} 
{\bf A}^{-1}{\bf B}=\left(\begin{array}{ccc}
\frac{x_1^{1-\alpha}}{\alpha S(\alpha)} & 0 & 0 \\
0 & \ddots & 0 \\
0 & 0 & \frac{x_d^{1-\alpha}}{\alpha S(\alpha)} 
\end{array} 
\right) \left( \begin{array}{c} x_1^{\alpha} \\ \vdots \\ x_{D-1}^{\alpha} \end{array} \right)=
\left(
\begin{array}{c}
\frac{x_1}{\alpha S(\alpha)}\\\vdots \\ \frac{x_{D-1}}{\alpha S(\alpha)}
\end{array} \right).
\end{eqnarray*} \\
Then the multiplication ${\bf C}^T{\bf A}^{-1}{\bf B}$ is 
\begin{eqnarray*} 
{\bf C}^T{\bf A}^{-1}{\bf B} &=& \left( \begin{array}{ccc} 
\alpha x_1^{\alpha-1}-\alpha x_D^{\alpha-1}, & \cdots, & \alpha x_{D-1}^{\alpha-1}-\alpha x_D^{\alpha-1} 
\end{array} \right)
\left(
\begin{array}{c}
\frac{x_1}{\alpha S(\alpha)}\\\vdots \\ \frac{x_{D-1}}{\alpha S(\alpha)}
\end{array} \right)  \\
&=& \frac{\sum_{i=1}^{D-1}x_i^{\alpha}}{S(\alpha)}-\frac{\alpha x_D^{\alpha-1}}{S(\alpha)}\left(\frac{x_1}{\alpha}+\cdots+\frac{x_{D-1}}{\alpha} \right).
\end{eqnarray*}
So we end up with 
\begin{eqnarray*} 1-{\bf C}^T{\bf A}^{-1}{\bf B} &=& \frac{S(\alpha)-\left(S(\alpha)-x_D^{\alpha}\right)}{S(\alpha)}+\frac{\alpha x_D^{\alpha-1}}{S(\alpha)}\sum_{i=1}^{D-1}\frac{x_i}{\alpha}=
\frac{x_D^{\alpha}+\alpha x_D^{\alpha-1}\sum_{i=1}^{D-1}\frac{x_i}{\alpha}}{S(\alpha)} \\
&=& \frac{x_D^{\alpha-1}\left(x_D+\alpha \sum_{i=1}^{D-1}\frac{x_i}{\alpha}\right)}{S(\alpha)}=\frac{x_D^{\alpha-1}}{S\left(\alpha\right)}.
\end{eqnarray*}
Finally the Jacobian of (\ref{stayalpha}) takes the following form 
\begin{eqnarray*} 
\left|{\bf J}\right| &=& S^{D-1}(\alpha)\frac{\prod_{i=1}^d\alpha x_i^{\alpha-1}}{S^{D-1}(\alpha)^{-2(D-1)}}\frac{x_D^{\alpha-1}}{S(\alpha)}=
S^{-(D-1)-1}(\alpha)x_D^{\alpha-1}\prod_{i=1}^{D-1}\alpha x_i^{\alpha-1}  \\ 
&=& \alpha^d \prod_{i=1}^D\frac{x_i^{\alpha-1}}{\sum_{j=1}^Dx_j^{\alpha}}.
\end{eqnarray*} 

The Jacobian of the $\alpha$-transformation (\ref{alpha}) without the left multiplication by the Helmert sub-matrix ${\bf H}$ is simply the Jacobian of (\ref{stayalpha}) multiplied by $\frac{D^{D-1}}{\alpha^{D-1}}$
\begin{eqnarray*} 
\left|{\bf J}\right|=D^{D-1}\prod_{i=1}^D\frac{x_i^{\alpha-1}}{\sum_{j=1}^Dx_j^{\alpha}}
\end{eqnarray*}
The multiplication by the Helmert sub-matrix adds an extra term to the Jacobian, which is $\sqrt{D}$ and hence the Jacobian becomes.
\begin{eqnarray*} 
\left|{\bf J}\right|=D^{D-1+1/2}\prod_{i=1}^D\frac{x_i^{\alpha-1}}{\sum_{j=1}^Dx_j^{\alpha}}
\end{eqnarray*}

\subsubsection*{Proof of Lemma 3.2} 
We will prove the Lemma 3.2 for the case that $\alpha=1$ for convenience purposes. The way to map a point $\bf x$ from inside the simplex to a point $\bf y$ outside of $\mathbb{A}_\alpha ^{D-1}$, is given in Equation (\ref{foldout}). Suppose we have a point $\mathbf{w} \in \mathbb{Q}_1^{D-1}$. We should apply the $\alpha$-transformation (\ref{alpha}) first and then apply the folding transformation.So, excluding the Helmert sub-matrix $\bf H$ and by simplifying our notation, we can write (\ref{foldout}) as follows.

\begin{eqnarray*}
{\bf y} =\left(\frac{1}{w^*(\mathbf{x})}\right)^2\mathbf{w}(\mathbf{x}) = \left(\frac{1}{w^*}\right)^2\mathbf{w}.
\end{eqnarray*}

We will prove the extra term in the Jacobian appearing in Lemma 3.2. The component wise transformation can be expressed as 
\begin{eqnarray} \label{foldingtrans}
y_i=Z_i\left(\frac{1}{w^*}\right)^2w_i+\left(1-Z_i\right)\left(\frac{1}{w_D}\right)^2w_i,
\end{eqnarray}
where $w_i$ refers to the $i$-th component of ${\bf w}_{\alpha}\left({\bf x}\right)$ defined in Equation (\ref{alef}), and since $\alpha=1$, we have excluded the superscript $\alpha$.  Also, 
\begin{eqnarray*} 
w^*=\big\vert \min \left\lbrace w_1,\ldots,w_{D-1} \right\rbrace \big\vert \ \ \text{and} \ \
Z_i=\left\lbrace \begin{array}{cc}
1 & \text{if} \ w^* \neq w_D \\
0 & \text{if} \ w^*=w_D 
\end{array} \right\rbrace, \ \ \text{for} \ \ i=1,\ldots,{D-1}. 
\end{eqnarray*}
There are two cases to consider when calculating the Jacobian determinant of the transformation.
\begin{enumerate}
\item The first case is when $Z_i=1$ and the transformation is 
\begin{eqnarray*}
y_i=\left(\frac{1}{w^*}\right)^2w_i.
\end{eqnarray*}
There are two sub-cases to be specified.
\begin{enumerate}
\item $w^*=w_i$ where the derivatives are given by
\begin{eqnarray*}
\frac{\partial y_i}{\partial w_j}= 
\left\lbrace \begin{array}{cc}
-\frac{1}{w_i^2}  & i = j \\ 
0                 & i \neq j.
\end{array} \right\rbrace.
\end{eqnarray*}
\item $w^* \neq w_i$ where the derivatives are given by 
\begin{eqnarray*}
\frac{\partial y_i}{\partial w_i}=\left(\frac{1}{w^*}\right)^2 \ \text{and} \ \
\frac{\partial y_i}{\partial w_j} = \left\lbrace \begin{array}{cc}
\frac{-2}{\left(w^*\right)^3}w_i  & \text{if} \ \ w^*=w_j \\
0 & \text{if} \ \ w^* \neq  w_j \end{array} \right\rbrace
\end{eqnarray*}
\end{enumerate}
The Jacobian matrix is
\begin{eqnarray*}
\left[ \begin{array}{ccccc} 
\frac{\partial y_1}{\partial w_1} & \ldots & \frac{\partial y_1}{\partial w_i} & \ldots  & \frac{\partial y_1}{\partial w_{D-1}} \\
\vdots                            & \ddots &       \vdots                      &         & \vdots                         \\                
\frac{\partial y_i}{\partial w_1} & \ldots & \frac{\partial y_i}{\partial w_i} & \ldots  & \frac{\partial y_i}{\partial w_{D-1}} \\
\vdots                            &        &       \vdots                      & \ddots  &  \vdots        \\
\frac{\partial y_{D-1}}{\partial w_1} & \ldots & \frac{\partial y_d}{\partial w_i} & \ldots  & \frac{\partial y_d}{\partial w_d} \\
\end{array}  \right] =
\left[ \begin{array}{cccccc} 
\frac{1}{\left(w^*\right)^2}   & 0                    & \ldots                & \ldots                & \ldots & 0       \\
0                  & \ddots   & 0                    & \ldots                & \vdots                & \vdots           \\
\vdots             & \ddots   & \frac{1}{\left(w^*\right)^2}     & \frac{-2}{\left(w^*\right)^3}  0    & \vdots           \\
\vdots             & \vdots   & 0                    & \ddots                & 0                     & \vdots           \\
\vdots             & 0        & \ldots               & \ddots                & \ddots                & 0                \\
0                  & \ldots   & \ldots               & \ldots                & 0                     & \frac{1}{\left(w^*\right)^2} \\
\end{array}  \right]
\end{eqnarray*}
and hence, the determinant is equal to 
\begin{eqnarray*}
\left|J\right|=\left(\frac{1}{w^*}\right)^{2(D-1)}.
\end{eqnarray*}
\item The second case is when $Z_i=0$ and the transformation is 
\begin{eqnarray*}
y_i=\frac{1}{w^2_D}w_i.
\end{eqnarray*}
The derivatives are now given by
\begin{eqnarray*}
\frac{\partial y_i}{\partial w_j}=\left\lbrace \begin{array}{cc}
\frac{1}{w^2_D}+\frac{2}{w^3_D}w_i & i=j \\ 
\frac{2}{w^3_D}w_i & i \neq j \end{array} \right\rbrace
\end{eqnarray*}
Note that the sign for the derivative with respect to $w_D$ is positive because
\begin{eqnarray*}
y_i &=& \frac{1}{w^2_D}w_i=\frac{w_i}{\left(-\sum_{j=1}^dw_j\right)^2}=\frac{w_i}{\left(\sum_{j=1}^dw_j\right)^2}, \ \ \text{thus} \\
\frac{\partial y_i}{\partial w_D} &=& -2\frac{w_i}{\left(\sum_{j=1}^dw_j\right)^3}=-2\frac{w_i}{\left(-w_D\right)^3}=
2\frac{w_i}{w^3_D}
\end{eqnarray*}
The Jacobian matrix in this case can be written as
\begin{eqnarray*}
\left[ \begin{array}{ccccc} 
\frac{\partial y_1}{\partial w_1} & \ldots & \frac{\partial y_1}{\partial w_i} & \ldots  & \frac{\partial y_1}{\partial w_{D-1}} \\
\vdots                            & \ddots &       \vdots                      &         & \vdots                         \\                
\frac{\partial y_i}{\partial w_1} & \ldots & \frac{\partial y_i}{\partial w_i} & \ldots  & \frac{\partial y_i}{\partial w_{D-1}} \\
\vdots                            &        &       \vdots                      & \ddots  &  \vdots        \\
\frac{\partial y_{D-1}}{\partial w_1} & \ldots & \frac{\partial y_d}{\partial w_i} & \ldots  & \frac{\partial y_{D-1}}{\partial w_{D-1}} \\
\end{array}  \right] =
\left[ \begin{array}{ccccc} 
\frac{1}{w^2_D}+\frac{2w_1}{w^3_D} & +\frac{2w_1}{w^3_D}  & \ldots        & \ldots & +\frac{2w_1}{w^3_D}  \\ 
\vdots                             & \ddots               & \ldots        & \vdots & \vdots               \\                
\frac{2w_i}{w^3_D}                 & \vdots               & \frac{1}{w^2_D}+\frac{2w_i}{w^3_D}  & \vdots & +\frac{2w_i}{w^3_D}  \\
\vdots                             & \vdots               & \ldots                              & \ddots &  \vdots             \\
\frac{2w_d}{w^3_D}                 & \ldots               & \ldots                 & \frac{2w_d}{w^3_D} & \frac{1}{w^2_D}+\frac{2w_{D-1}}{w^3_D} \\
\end{array}  \right]
\end{eqnarray*}
The determinant of such matrices is given by (Mardia, Kent, \& Bibby, 1979) 
\begin{eqnarray} \label{jack}
\left|J\right|=\left|{\bf A}+{\bf BC}^T\right|=\left|{\bf A} \right|\left(1+{\bf C}^T{\bf A}^{-1}{\bf B} \right) 
\end{eqnarray}
where 
\begin{eqnarray*}
{\bf A}=\text{diag}\left(\frac{1}{w^2_D},\ldots,\frac{1}{w^2_D}\right), \ {\bf B}=2\left(w_1,\ldots, w_{D-1} \right)^T \ \text{and} \ {\bf C}=\left(w_D^{-3},\ldots, w_D^{-3} \right)^T.
\end{eqnarray*}
\begin{eqnarray*}
\left|J\right| &=& \left(\frac{1}{w^2_D}\right)^{D-1} \left \lbrace 1+
 2\left(w_D^{-3},\ldots, w_D^{-3} \right)
\left[ \begin{array}{ccc} 
w^2_D   & {\bf 0} & {\bf 0} \\
{\bf 0} & \ddots  & {\bf 0} \\
{\bf 0} & {\bf 0} & w^2_D     
\end{array}  \right]
\left( \begin{array}{c} 
w_1   \\
\vdots \\
w_d 
\end{array}\right)  \right\rbrace \\ 
&=& \left(\frac{1}{w^2_D}\right)^{D-1} \left[ 1 + 2\left(w_D^{-1},\ldots, w_D^{-1} \right)
\left( \begin{array}{c} 
w_1   \\
\vdots \\
w_d
\end{array}\right) \right]  \\
&=& \left(\frac{1}{w^2_D}\right)^{D-1} \left[ 1+ 2\frac{\sum_{j=1}^{D-1}w_j}{w_D} \right] = \left(\frac{1}{w^2_D}\right)^{D-1} \left(1+ 2\frac{\sum_{j=1}^{D-1}w_j}{-\sum_{j=1}^{D-1}w_j} \right).
\end{eqnarray*} 
Finally, (\ref{jack}) becomes 
\begin{eqnarray*}
\left|J\right|=\left|\left(\frac{1}{w^2_D}\right)^{D-1}\left(1-2\right)\right|=\left(\frac{1}{w_D}\right)^{2(D-1)}.
\end{eqnarray*}
\end{enumerate}

\end{appendix}

\end{document}